\documentclass{article} 
\usepackage{iclr2027_conference,times}

\usepackage{amsmath,amsfonts,bm}

\def\eqref#1{equation~\ref{#1}}

\def\1{\bm{1}}

\DeclareMathAlphabet{\mathsfit}{\encodingdefault}{\sfdefault}{m}{sl}
\SetMathAlphabet{\mathsfit}{bold}{\encodingdefault}{\sfdefault}{bx}{n}

\usepackage{hyperref}
\usepackage{url}
\usepackage{amssymb} 
\usepackage{amsthm}
\usepackage{algorithm}
\usepackage{algorithmic}
\usepackage{booktabs}
\usepackage{graphicx} 
\usepackage{multirow}
\usepackage{bbm}
\usepackage{wrapfig}
\usepackage[table]{xcolor}
\usepackage{xspace}
\usepackage{enumitem}
\definecolor{BestRed}{RGB}{190,30,45}
\definecolor{SecondBlue}{RGB}{35,85,160}
\definecolor{RankGray}{gray}{0.90}
\definecolor{OursTint}{RGB}{240,245,252}
\newcommand{\best}[1]{\textbf{\textcolor{BestRed}{#1}}}
\newcommand{\secondb}[1]{\underline{\textcolor{SecondBlue}{#1}}}
\newcommand{\method}{\textsc{DyG-WM}\xspace}

\title{Dynamic Graph World Modelling: Learning Future Dynamics from Temporal Context}

\author{
\textbf{Bo Li$^{1,2}$, Yue Yang$^{2}$, Yujin Huang$^{3}$, Siao Liu$^{4}$,} 
\textbf{David Lemphers$^{2}$},\\\textbf{Sara E Rodriguez Gomez$^{2}$, Luke Borgnolo$^{2}$,}
\textbf{Can Wang$^{1}$, Xin Zheng$^{5}$\thanks{Corresponding author}} \\
{\normalfont
$^{1}$School of Information and Communication Technology, Griffith University, Australia} \\
{\normalfont $^{2}$Maincode, Australia} \\
{\normalfont
$^{3}$School of Computing and Information Systems, University of Melbourne, Australia} \\
{\normalfont
$^{4}$Future Science and Engineering College, Soochow University, China} \\
{\normalfont
$^{5}$School of Computing Technologies, RMIT University, Australia}\\
{\normalfont\texttt{bo.li6@griffithuni.edu.au,\quad xin.zheng2@rmit.edu.au}}
}

\iclrfinalcopy 

\begin{document}

\maketitle
\fancyhead{}
\begin{abstract}
Dynamic graph learning models how entities and their relations evolve from historical observations to predict future graph states. Existing methods do not fully address dynamic graph world modelling, which requires learning transitions from temporal context to future latent graph states. This is difficult for three reasons: nodes and relations rely on different historical evidence, structural evolution is path-dependent, and future snapshots provide sparse and realization-specific targets. We propose \textbf{\textsc{\underline{Dy}}}namic \textbf{\textsc{\underline{G}}}raph \textbf{\textsc{\underline{W}}}orld \textbf{\textsc{\underline{M}}}odelling (\textbf{\method}), the first framework to jointly predict future node and relation representations from temporal context by answering three questions: \emph{what has evolved, how it evolved, and what will evolve next}. To capture (a) \emph{what has evolved}, a historical node-relation context encoder summarizes node-neighborhood trajectories and pair-specific histories. To characterize (b) \emph{how it evolved},  a node-relation path-signature encoder captures ordered structural changes. To predict (c) \emph{what will evolve next}, a dual-level joint-embedding objective predicts target-snapshot node and relation representations without reconstructing the raw graph. Experiments on 12 link-prediction and two node-classification datasets demonstrate the effectiveness of \method across both tasks. Our code available at \nolinkurl{https://github.com/HyoukaLI/DyG-WM}
\end{abstract}
\section{Introduction}
Many real-world systems change through the evolution of both entities and their relationships. Dynamic graphs model these systems as sequences of graph-structured states, in which nodes represent entities and time-varying edges record their interactions~\citep{gravina2024long, feng2025comprehensive, skarding2021foundations}. They have been widely used to study social networks~\citep{kumar2019predicting, zheng2026dp}, recommender systems~\citep{zhang2026modeling, wang2026embedding}, traffic systems~\citep{zou2026meta, weng2023decomposition}, financial transactions~\citep{song2023towards, yuan2025dynamic}, and scientific discovery~\citep{bai2024dynamic}. By modelling structural dependencies within graph states and temporal dependencies across states, dynamic graph neural networks (DGNNs) can do temporal link prediction and node attribute prediction~\citep{kumar2019predicting,rossi2020temporal,yu2023towards,li2026dyg}. Although current models perform well on individual prediction tasks, real-world applications require more than forecasting a single edge or node label. Dynamic graph learning aims to model how the graph state evolves from historical observations so that future states can be predicted and used for planning and decision-making.

\begin{wrapfigure}{r}{0.6\textwidth}
    \vspace{-20pt}
    \centering
    \includegraphics[width=0.6\textwidth]{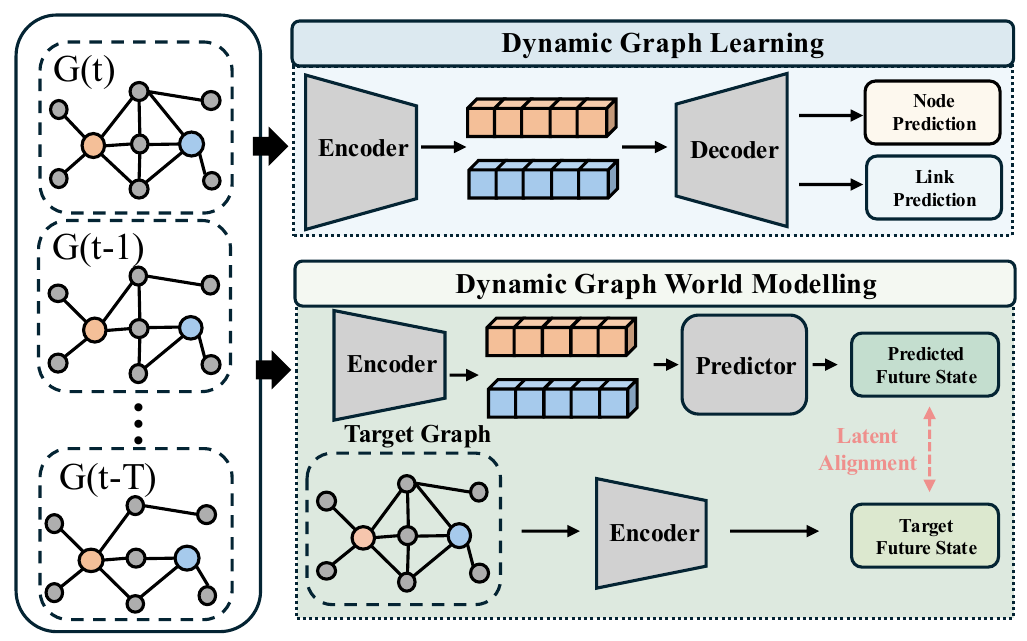}
    \vspace{-15pt}
    \caption{Comparison of dynamic graph learning and dynamic graph world modelling. Dynamic graph learning maps historical graphs to task-specific outputs, whereas dynamic graph world modelling predicts future latent state.}
    \label{fig:fig1}
    \vspace{-5pt}
\end{wrapfigure}

This motivates us to formulate the problem of \textbf{dynamic graph world modelling}, which extends the general world-modelling paradigm to evolving relational systems. As illustrated in Figure~\ref{fig:fig1}, historical graph observations are encoded into compact latent states, and the temporal transitions between these states are learned to predict future system dynamics. From this perspective, existing dynamic graph learning methods can be broadly divided into two groups, neither of which fully addresses this problem. \emph{(1) Task-oriented temporal graph representation learning} trains DGNNs with node labels or observed interactions to predict specific outcomes, such as future links or node attributes~\citep{kumar2019predicting,rossi2020temporal,yu2023towards,li2026dyg}. Although effective on individual tasks, such supervision is often sparse, delayed, or costly to obtain and binds the learned representations to particular tasks and evolving environments. \emph{(2) Self-supervised dynamic graph representation learning} reduces this dependence through contrastive, reconstruction-based, or latent-predictive objectives. Contrastive methods align temporal views, sampled subgraphs, or interacting nodes~\citep{wang2021tcl,xu2023cldg}. However, they mainly enforce invariance across constructed views, and their effectiveness depends on whether the selected positive-negative pairs preserve temporal semantics. Reconstruction and masking methods recover missing edges, attributes, or temporal observations~\citep{gao2025dvgmae,he2025maskdgnn}, but encourage the model to focus on low-level details of the observed graph rather than its predictable dynamics. Dynamic graphs have a natural temporal prediction structure: past graph states provide the context, while future graph states serve as prediction targets. Predictive latent representation learning is therefore well suited to dynamic graph world modelling because temporal progression naturally separates context from target. The goal is to predict, from historical states, what the graph will become in representation space rather than reconstruct every observed interaction.

However, learning a dynamic graph world model involves three challenges. \textbf{C1: Multi-level historical dependency.} Nodes and relations evolve together but draw on different historical evidence. Node evolution depends on the trajectory of a node and its neighborhood, whereas relation evolution also depends on pair-specific structural histories that endpoint representations alone cannot recover. \textbf{C2: Path-dependent structural evolution.} Similar current graph states can result from different sequences of structural changes and may therefore lead to different future dynamics. A world model must retain both the resulting state and the ordered path that produced it. \textbf{C3: Sparse and realization-specific future targets.} A future graph snapshot records only the node states and interactions observed within a given time window, offering an incomplete view of the underlying dynamics. Predicting these outcomes separately or reconstructing the snapshot exactly may lead the model to fit a single observed realization rather than capture how nodes and relations evolve more generally. A dynamic graph world model therefore needs a compact future-state target that goes beyond individual node labels and edge observations.

To address these challenges, we introduce \textbf{\method}, a \textbf{\textsc{\underline{Dy}}}namic \textbf{\textsc{\underline{G}}}raph \textbf{\textsc{\underline{W}}}orld \textbf{\textsc{\underline{M}}}odelling method that learns future graph dynamics directly in the latent space through predictive representation learning guided by temporal context. 
Our proposed \method is designed and organized around three questions: \emph{what has evolved, how it evolved, and what will evolve next}. To capture \emph{what has evolved}, we develop \textbf{(a) a historical node-relation context encoder} that summarizes the trajectories of nodes and their neighborhoods and builds pair-specific structural contexts. This provides the different forms of historical evidence described in C1. To characterize \emph{how the graph evolved}, we introduce \textbf{(b) a node-relation path-signature encoder} that captures the order of structural changes at both levels. It distinguishes trajectories that reach similar current states through different transition orders, addressing the path dependence described in C2. Finally, to predict \emph{what will evolve next}, we design \textbf{(c) a dual-level joint-embedding predictive objective} that uses historical contexts and path signatures to predict node and relation representations from the target snapshot. By learning future dynamics in representation space instead of reconstructing a single future graph realization, the objective addresses C3.
Our main contributions are as follows:
\begin{itemize}[leftmargin=1.5em]
\item To our knowledge, \method is the first framework to formulate dynamic graph world modelling as the joint prediction of future node and relation representations. It treats relations as temporally evolving latent states rather than only as message-passing channels or endpoint-based scores. Using strictly historical context, \method learns future dynamics directly in representation space.

\item To account for path dependence in dynamic graph evolution, we combine a historical node-relation context encoder with a node-relation path-signature encoder. The two encoders represent accumulated states and ordered structural changes, respectively. A dual-level joint-embedding predictive objective uses these encodings to predict future node and relation states without relying on augmented-view contrast or raw-graph reconstruction for self-supervision.

\item Across 12 dynamic link-prediction datasets and two dynamic node-classification datasets, \method achieves the best average rank in every evaluated link-prediction setting. It also obtains the best Macro-F1 and Micro-F1 scores across all label ratios. Ablation and latent-alignment analyses further confirm the effectiveness of the individual components.
\end{itemize}

\section{Related Works}
\noindent\textbf{Dynamic Graph Neural Networks.} Dynamic graph neural networks (DGNNs) are commonly divided into discrete-time models, which operate on graph snapshots, and continuous-time models, which process timestamped interactions directly. Representative methods capture graph evolution through temporal attention~\citep{sankar2020dysat}, recurrent updates~\citep{pareja2020evolvegcn}, memory mechanisms~\citep{rossi2020temporal}, temporal walks~\citep{wang2021inductive}, MLP-based temporal mixing~\citep{cong2023we}, Transformer-based sequence modeling~\citep{yu2023towards}, and state-space models~\citep{li2026dyg}. Recent self-supervised DGNNs learn from temporal contrast, masked reconstruction, or future latent prediction \citep{chen2025signn,gao2025dvgmae,he2025maskdgnn,zhang2026scalable}. However, most remain node-centric, while pair-level information is mainly used for link prediction \citep{wang2021inductive,wang2021tcl,yu2023towards,tian2024freedyg}. \method instead models and predicts future representations at both node and relation levels.

\noindent\textbf{Graph World Model.} World models learn latent states and transition functions to predict future dynamics \citep{ha2018world}. Graph world models extend this idea to structured environments by representing states as entities and relations, with a graph abstraction and a relational transition model \citep{battaglia2018relational,liu2026graph}. Existing object-centric models use graph neural networks to predict interactions and future object states, including C-SWM~\citep{kipf2020cswm}, G-SWM~\citep{lin2020gswm}, CWM~\citep{li2020cwm}, VGPL~\citep{li2020vgpl}, and other graph-based world models~\citep{feng2025graph,zhang2021world,zhu2022value}. These methods mainly target action-conditioned physical systems, where relations support object-state updates rather than being predicted themselves. \method instead models general dynamic graphs by constructing historical node and relation contexts and jointly predicting their future representations. It can therefore be viewed as an action-free relational transition model rather than a full planning system.

\section{Methodology}
\subsection{Preliminary}
\paragraph{Notation.} Let $\mathcal V$ be a global node set with $N=|\mathcal V|$. A dynamic graph is a time-ordered interaction stream
$\mathcal E=\bigl((u_i,v_i,\tau_i,\mathbf e_i)\bigr)_{i=1}^{M}$ with $\tau_i\leq\tau_{i+1}$, where ordered node pairs may repeat. We partition $\mathcal E$ into $T$ consecutive subsequences $\{\mathcal E_t\}_{t=1}^{T}$ with roughly equal event counts, keeping equal-timestamp interactions in the same bin.
Each bin has two aligned views. The snapshot view is $\widetilde G_t=(\mathcal V,A_t,X_t,\mathbf m_t)$, where $A_t\in\{0,1\}^{N\times N}$ is the de-duplicated, symmetrized message-passing adjacency, $X_t\in\mathbb R^{N\times F}$ is the node-attribute matrix, and $\mathbf m_t\in\{0,1\}^{N}$ is the active-node mask. We define $\mathcal V_t=\{u\in\mathcal V:m_u^t=1\}$ and write $\mathcal N_t(u)$ for the neighbors of $u$ in $A_t$. Both $A_t$ and $X_t$ remain globally indexed. Time-varying components of $X_t$ summarize events through bin $t$, with global preprocessing statistics computed from the training split only.
The event view $\mathcal E_t$ preserves directed interactions, repetitions, exact timestamps, and link features. For query time $\tau$, the strict history is
$\mathcal H_{<\tau}=\bigl((u_i,v_i,\tau_i,\mathbf e_i):\tau_i<\tau\bigr)$. Equal-timestamp events are simultaneous and excluded from one another's histories. Superscript $t$ denotes bin-level indexing, whereas $\tau$ or $<\tau$ denotes exact-time dependence.

\paragraph{Problem Formalization.} For target bin $t$, let $\mathcal C_t=(\widetilde G_{t-W+1},\ldots,\widetilde G_{t-1})$ contain the preceding $W-1$ snapshot views and no event information from bin $t$. From $\mathcal C_t$, the model computes a node context $\mathbf c_u^t$, which, despite its index $t$, depends only on snapshots before the target bin. For query $(u,v,\tau)$ in bin $t$, the model combines $\mathcal C_t$ and $\mathcal H_{<\tau}$ to obtain the exact-time relation representation $\mathbf z_{uv}^{\tau}$. Dynamic link prediction scores an indexed candidate group $\mathcal Q(u,\tau)$ containing one observed interaction and $K$ sampled negatives; sampled destinations that coincide with the observed destination are retained rather than deduplicated.

Given the historical snapshot context $\mathcal C_t$ and strict event history $\mathcal H_{<\tau}$, we learn a shared predictive representation for node classification and dynamic link prediction, while keeping the target snapshot $\widetilde G_t$ hidden from the online branch.

\begin{figure}
    \centering
    \includegraphics[width=\textwidth]{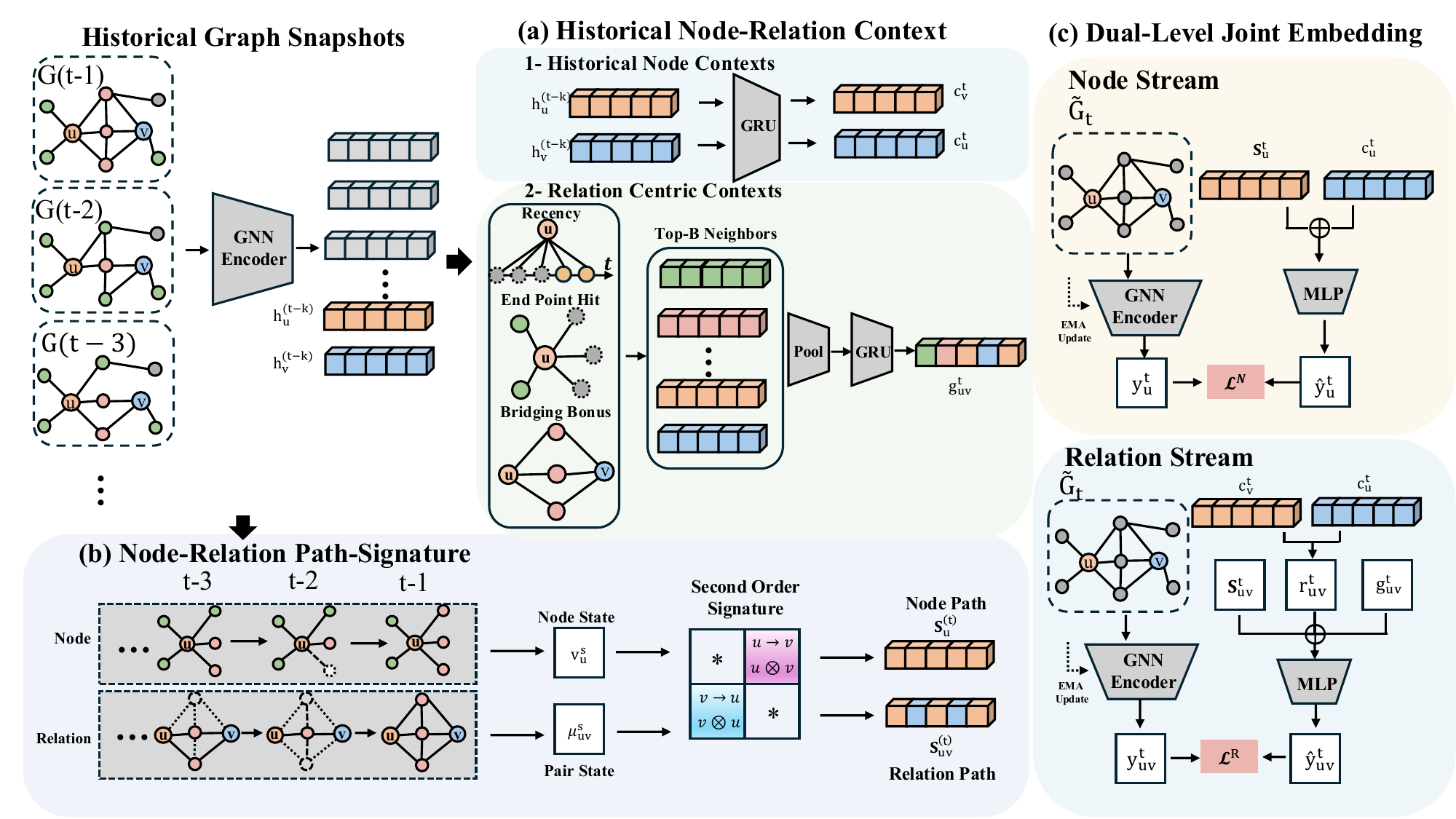}
    \vspace{-10pt}
    \caption{Overview of \method. (a) The historical node-relation context encoder summarizes the trajectories of nodes and their neighborhoods and constructs pair-specific relation contexts from historical snapshots. (b) The node-relation path-signature encoder uses second-order signatures to capture ordered structural changes at both levels. (c) The dual-level joint-embedding predictive objective predicts target-snapshot node and relation representations.}
    \label{fig:framework}
    \vspace{-10pt}
\end{figure}

\label{sec:method}
The overall framework of \method~is shown in Figure~\ref{fig:framework}. \method~consists of three main components. The historical node-relation context encoder summarizes node trajectories and constructs pair-specific relation contexts from preceding snapshots. The node-relation path-signature encoder models ordered structural changes at both the node and relation levels. The dual-level joint-embedding predictive objective predicts target-snapshot node and relation representations using an EMA target encoder. Given the historical snapshot context $\mathcal C_t$ and the strict event prefix $\mathcal H_{<\tau}$ for a query $(u,v,\tau)$, \method~constructs a relation representation and produces the interaction score $s_{uv}^{\tau}$. The following sections describe these components in detail.
\subsection{Historical Node-Relation Context Encoder}
\label{sec:historical-context}

Predicting future dynamics requires historical information at both the node and relation levels. Node prediction tracks how a node and its neighborhood evolve over time, while relation prediction also depends on pair-specific structural history that is not captured by the endpoint representations alone. We use a \underline{(a-1)} historical node context encoder for node-level evolution and a \underline{(a-2)} relation-centric context encoder for pair-specific structural history.

\noindent\textbf{Historical Node Context Encoder.}
For target bin $t$, let $\mathcal I_t=\{t-W+1,\ldots,t-1\}$ index the historical snapshots in $\mathcal C_t$. Each node representation is obtained from the online GNN as
\begin{equation}
\mathbf h_u^s=\operatorname{GNN}_{\theta}(\widetilde G_s)[u],\qquad s\in\mathcal I_t.
\label{eq:historical-snapshot-encoding}
\end{equation}
The latest representation $\mathbf h_u^{t-1}$ alone does not capture the earlier evolution of node $u$ or its neighborhood. We therefore compute the mean neighborhood representation
\begin{equation}
\overline{\mathbf h}_u^s=\frac{1}{\max\{1,|\mathcal N_s(u)|\}}\sum_{w\in\mathcal N_s(u)}\mathbf h_w^s,
\label{eq:historical-neighborhood}
\end{equation}
with $\overline{\mathbf h}_u^s=\mathbf 0$ when $\mathcal N_s(u)=\varnothing$.

The historical node context encoder processes the ordered node and neighborhood trajectories with separate recurrent encoders and combines them with the latest node representation through gated residual fusion:
\begin{equation}
\mathbf c_u^t=f_{\mathrm{NodeC}}\left(\mathbf h_u^{t-1},(\mathbf h_u^s)_{s\in\mathcal I_t},(\overline{\mathbf h}_u^s)_{s\in\mathcal I_t}\right).
\label{eq:historical-node-context}
\end{equation}
Thus, $\mathbf c_u^t$ summarizes the recent state and historical node-neighborhood evolution of $u$ using only snapshots $s<t$.

\noindent\textbf{Relation-Centric Context Encoder.}
For a queried pair $(u,v)$, we collect the endpoints and their historical neighbors over $\mathcal I_t$. The endpoints are retained, while the remaining candidates are ranked using recency and common-neighbor bridging information. We keep the top $B$ context nodes as $\mathcal K_{uv}^t$, where $B$ is the relation-context budget. Their representations are pooled at each historical snapshot and summarized temporally:
\begin{equation}
\mathbf g_{uv}^t=\operatorname{GRU}_{\mathrm{rel}}\left(\left(\frac{1}{|\mathcal K_{uv}^t|}\sum_{w\in\mathcal K_{uv}^t}\mathbf h_w^s\right)_{s\in\mathcal I_t}\right).
\label{eq:historical-relation-context}
\end{equation}
The resulting $\mathbf g_{uv}^t$ captures pair-specific structural history using only $\mathcal C_t$.

\subsection{Node-Relation Path-Signature Encoder}
\label{sec:path-signature}

Dynamic graph evolution is path-dependent: similar current states may arise from different historical trajectories and lead to different future dynamics. Although $\mathbf c_u^t$ and $\mathbf g_{uv}^t$ summarize the states in $\mathcal C_t$, recurrent summaries do not explicitly preserve the order and interactions of structural changes.
Node and relation prediction also require different trajectories. Node prediction depends on local neighborhood evolution, while relation prediction additionally depends on the joint evolution of the queried pair. We therefore use a node structural path-signature encoder and a relation structural path-signature encoder to model these ordered changes~\citep{chevyrev2025primer}.

\noindent\textbf{Second-Order Path Signatures.}
Let $\mathbf d_1,\ldots,\mathbf d_L\in\mathbb R^{d_s}$ be a sequence of projected path increments, where $L=W-1$. We represent the ordered path using the truncated second-order signature
\begin{equation}
\operatorname{Sig}_2(\mathbf d_{1:L})=\left[\sum_{i=1}^{L}\mathbf d_i\ \Big\Vert\ \operatorname{vec}\left(\sum_{1\leq i<j\leq L}\mathbf d_i\otimes\mathbf d_j+\frac{1}{2}\sum_{i=1}^{L}\mathbf d_i\otimes\mathbf d_i\right)\right]
\label{eq:truncated-signature}
\end{equation}
The first component summarizes cumulative change, while the second captures ordered interactions between structural changes through the terms with $i<j$.

\noindent\textbf{Node Structural Paths.}
For each $s\in\mathcal I_t$, we construct a structural descriptor $\boldsymbol{\nu}_u^s\in\mathbb R^{d_N}$ from the normalized degree of $u$, its signed degree change, the proportions of arriving, disappearing, and persistent neighbors, and an activity indicator. For the first historical snapshot $s_0=t-W+1$, transition-dependent entries are set to zero; subsequent transitions use snapshots $s-1$ and $s$.

We append the normalized elapsed time $\overline{\Delta t}_s$, project each descriptor to the path space, and apply the second-order signature:
\begin{equation}
\mathbf S_u^t=\phi_N\left(\operatorname{Sig}_2\left(\left(\operatorname{Proj}_N[\overline{\Delta t}_s\Vert\boldsymbol{\nu}_u^s]\right)_{s\in\mathcal I_t}\right)\right).
\label{eq:node-path-signature}
\end{equation}
Here, $\phi_N$ maps the truncated signature to the model dimension. While $\mathbf c_u^t$ summarizes the historical state of $u$, $\mathbf S_u^t$ captures its ordered structural evolution.

\noindent\textbf{Relation Structural Paths.}
For each queried pair $(u,v)$ and historical snapshot $s$, we construct a pair state $\boldsymbol{\mu}_{uv}^s\in\mathbb R^{d_R}$ from direct connectivity, normalized endpoint degrees, normalized common-neighbor count, neighborhood Jaccard overlap, endpoint activity, and coverage of the selected context nodes $\mathcal K_{uv}^t$. Endpoint-specific channels remain ordered to preserve the roles of $u$ and $v$.

We describe the relation trajectory by changes in these pair states:
\begin{equation}
\Delta\boldsymbol{\mu}_{uv}^s=\begin{cases}\boldsymbol{\mu}_{uv}^s, & s=t-W+1,\\ \boldsymbol{\mu}_{uv}^s-\boldsymbol{\mu}_{uv}^{s-1}, & s>t-W+1.\end{cases}
\label{eq:relation-state-increment}
\end{equation}

After appending the normalized elapsed time and projecting each increment to the path space, we apply the second-order signature:
\begin{equation}
\mathbf S_{uv}^t=\phi_R\left(\operatorname{Sig}_2\left(\left(\operatorname{Proj}_R[\overline{\Delta t}_s\Vert\Delta\boldsymbol{\mu}_{uv}^s]\right)_{s\in\mathcal I_t}\right)\right).
\label{eq:relation-path-signature}
\end{equation}
Here, $\mathbf S_{uv}^t$ captures the ordered structural evolution of $(u,v)$ using only $\mathcal C_t$. Together, $\mathbf S_u^t$ and $\mathbf S_{uv}^t$ describe ordered node-level and pair-level structural changes that complement $\mathbf c_u^t$ and $\mathbf g_{uv}^t$.
\subsection{Dual-Level Joint-Embedding Predictive Objective}
\label{sec:jepa-objective}
Future observations in dynamic graphs are sparse and variable, so directly predicting individual links or node labels can overfit to a particular realization of the future graph. We instead predict future latent states, which capture structural changes without reconstructing the exact target snapshot. Since these changes occur at both node and pair levels, we use a dual-level joint-embedding predictive objective for future node and relation states. We therefore predict future latent representations at both levels instead of reconstructing target adjacency or raw attributes.

\noindent\textbf{Online and Target Branches.}
We use $\operatorname{GNN}_{\theta}$ and $g_{\theta}$ as the online snapshot and relation encoders, with EMA target counterparts $\operatorname{GNN}_{\bar{\theta}}$ and $g_{\bar{\theta}}$. The online branch uses only historical information, while the target branch encodes $\widetilde G_t$ during training. The node target is
$\mathbf y_u^t=\operatorname{sg}\left(\operatorname{GNN}_{\bar{\theta}}(\widetilde G_t)[u]\right), u\in\mathcal V_t$,
where $\operatorname{sg}(\cdot)$ denotes stop-gradient. The target parameters receive no gradients and are updated by EMA:
$\bar{\theta}\leftarrow\eta_{\mathrm{ema}}\bar{\theta}+(1-\eta_{\mathrm{ema}})\theta$.

\noindent\textbf{Node-Level Predictive Objective.}
The node predictor uses the historical context, path signature, and prediction horizon:
$\widehat{\mathbf y}_u^t=p_N\left([\mathbf c_u^t\Vert\mathbf S_u^t\Vert\mathbf e_{\Delta_t}]\right)$,
where $\mathbf e_{\Delta_t}$ encodes the prediction horizon. Prediction and target are compared after $\ell_2$ normalization:
\begin{equation}
D(\widehat{\mathbf y},\mathbf y)=\left\|\frac{\widehat{\mathbf y}}{\|\widehat{\mathbf y}\|_2}-\operatorname{sg}\left(\frac{\mathbf y}{\|\mathbf y\|_2}\right)\right\|_2^2.
\label{eq:normalized-latent-distance}
\end{equation}
The node loss is
$\mathcal L^N=\frac{1}{|\mathcal U_t|}\sum_{u\in\mathcal U_t}D(\widehat{\mathbf y}_u^t,\mathbf y_u^t)$,
where $\mathcal U_t=\mathcal V_t$ for node prediction and contains the observed-interaction endpoints in the current batch for link prediction.

\noindent\textbf{Relation-Level Predictive Objective.}
We use the ordered endpoint transformation $\chi(\mathbf a,\mathbf b)=[\mathbf a\Vert\mathbf b\Vert\mathbf a\odot\mathbf b\Vert\mathbf a-\mathbf b]$. The historical relation representation is
\begin{equation}
\mathbf q_{uv}^t=\phi_{\mathrm{fuse}}\left([g_{\theta}(\chi(\mathbf c_u^t,\mathbf c_v^t))\Vert\mathbf g_{uv}^t\Vert\mathbf S_{uv}^t]\right).
\label{eq:historical-relation-representation}
\end{equation}
Its future latent is predicted as
$\widehat{\mathbf y}_{uv}^t=p_R\left([\mathbf q_{uv}^t\Vert\mathbf e_{\Delta_t}]\right)$,
with target
$\mathbf y_{uv}^t=\operatorname{sg}\left(g_{\bar{\theta}}(\chi(\mathbf y_u^t,\mathbf y_v^t))\right)$.
Explicit target-snapshot pair statistics, such as adjacency and common-neighbor counts, are not provided to $g_{\bar{\theta}}$. For observed relation pairs $\mathcal R_t^+$,
$\mathcal L^R=\frac{1}{|\mathcal R_t^+|}\sum_{(u,v)\in\mathcal R_t^+}D(\widehat{\mathbf y}_{uv}^t,\mathbf y_{uv}^t)$.
Sampled negatives are used only by the downstream link objective.
\noindent\textbf{Joint Objective.}
We apply variance and covariance regularization to the predicted node and relation latents to reduce collapse. The resulting objective is
\begin{equation}
\mathcal L_{\mathrm{JEPA}}=\lambda_N\mathcal L^N+\lambda_R\mathcal L^R+\lambda_{\mathrm{var}}\left(\mathcal L_{\mathrm{var}}^N+\mathcal L_{\mathrm{var}}^R\right)+\lambda_{\mathrm{cov}}\left(\mathcal L_{\mathrm{cov}}^N+\mathcal L_{\mathrm{cov}}^R\right).
\label{eq:dual-level-jepa-objective}
\end{equation}
For node prediction, we additionally use in-batch node discrimination. For link prediction,
$$\mathcal L_{\mathrm{total}}=\mathcal L_{\mathrm{JEPA}}+\lambda_{\mathrm{link}}\mathcal L_{\mathrm{link}}+\lambda_{\mathrm{rank}}\mathcal L_{\mathrm{rank}},$$
where $\mathcal L_{\mathrm{link}}$ is the binary candidate loss and $\mathcal L_{\mathrm{rank}}$ ranks the observed destination against sampled candidates.

\noindent\textbf{Strict Event-Prefix Encoder.}
The bin-level representation $\mathbf q_{uv}^t$ is fixed within target bin $t$ and does not use events from that bin. To include information available before the exact query time, we encode the strict prefix $\mathcal H_{<\tau}$. The event-history encoder uses statistics of the directed pair and its endpoints, together with historical link-feature profiles when available:
\begin{equation}
\mathbf o_{uv}^{<\tau}=\operatorname{EventHistory}(u,v;\mathcal H_{<\tau}).
\label{eq:event-history-encoder}
\end{equation}
All statistics are computed strictly before $\tau$, so equal-timestamp events are excluded.

\noindent\textbf{Training and Inference.}
The target snapshot $\widetilde G_t$ is used only to construct $\mathbf y_u^t$ and $\mathbf y_{uv}^t$. For query $(u,v,\tau)$,
$\mathbf z_{uv}^{\tau}=\phi_{\tau}\left([\mathbf q_{uv}^t\Vert\mathbf o_{uv}^{<\tau}]\right), s_{uv}^{\tau}=h_{\psi}(\mathbf z_{uv}^{\tau})$.
Inference therefore uses only $\mathcal C_t$ and $\mathcal H_{<\tau}$; the target branch and predictive losses are used only during training.
\begin{table*}[h]
\vspace{-10pt}
\centering
\caption{\textbf{Transductive} AUC-ROC for dynamic link prediction with random (\emph{rnd}), historical (\emph{hist}), and inductive (\emph{ind}) negative edge sampling. \best{Bold red} and \secondb{underlined blue} emphasize best and 2nd-best results.}
\label{tab:auc_main}
\resizebox{\textwidth}{!}{%
\begin{tabular}{llccccccccccc>{\columncolor{OursTint}}c}
\toprule
NSS & Datasets & JODIE & DyRep & TGAT & TGN & CAWN & TCL & GraphMixer & DyGFormer & CLDG & MaskDGNN & DVGMAE & \method~ \\
\midrule
\multirow{13}{*}{\emph{rnd}} & Wikipedia & 96.55{\scriptsize $\pm$ 0.19} & 94.93{\scriptsize $\pm$ 0.50} & 92.66{\scriptsize $\pm$ 0.43} & 98.46{\scriptsize $\pm$ 0.05} & 98.60{\scriptsize $\pm$ 0.01} & 97.51{\scriptsize $\pm$ 0.09} & 97.12{\scriptsize $\pm$ 0.06} & \best{98.98{\scriptsize $\pm$ 0.01}} & 89.50{\scriptsize $\pm$ 0.18} & 86.71{\scriptsize $\pm$ 0.53} & 83.96{\scriptsize $\pm$ 0.55} & \secondb{98.94{\scriptsize $\pm$ 0.02}} \\
 & Reddit & 98.04{\scriptsize $\pm$ 0.04} & 97.69{\scriptsize $\pm$ 0.51} & 97.97{\scriptsize $\pm$ 0.02} & 97.75{\scriptsize $\pm$ 0.38} & \secondb{99.10{\scriptsize $\pm$ 0.01}} & 97.91{\scriptsize $\pm$ 0.04} & 98.03{\scriptsize $\pm$ 0.02} & 99.09{\scriptsize $\pm$ 0.00} & 96.75{\scriptsize $\pm$ 0.01} & 96.73{\scriptsize $\pm$ 0.02} & 95.72{\scriptsize $\pm$ 0.07} & \best{99.20{\scriptsize $\pm$ 0.01}} \\
 & MOOC & 83.16{\scriptsize $\pm$ 2.19} & 79.33{\scriptsize $\pm$ 4.02} & 86.01{\scriptsize $\pm$ 0.06} & \best{94.19{\scriptsize $\pm$ 0.73}} & 82.05{\scriptsize $\pm$ 0.28} & 85.58{\scriptsize $\pm$ 0.39} & 84.98{\scriptsize $\pm$ 0.13} & \secondb{89.80{\scriptsize $\pm$ 0.23}} & 68.17{\scriptsize $\pm$ 1.30} & 74.97{\scriptsize $\pm$ 0.20} & 71.02{\scriptsize $\pm$ 0.89} & 82.50{\scriptsize $\pm$ 0.39} \\
 & LastFM & 70.94{\scriptsize $\pm$ 0.72} & 71.92{\scriptsize $\pm$ 0.84} & 71.54{\scriptsize $\pm$ 0.35} & 62.38{\scriptsize $\pm$ 2.16} & 85.93{\scriptsize $\pm$ 0.11} & 71.10{\scriptsize $\pm$ 0.32} & 73.59{\scriptsize $\pm$ 0.11} & \best{93.15{\scriptsize $\pm$ 0.02}} & 64.49{\scriptsize $\pm$ 0.69} & 65.48{\scriptsize $\pm$ 0.66} & 63.00{\scriptsize $\pm$ 0.21} & \secondb{92.61{\scriptsize $\pm$ 0.06}} \\
 & Enron & 87.29{\scriptsize $\pm$ 1.84} & 85.60{\scriptsize $\pm$ 1.98} & 74.41{\scriptsize $\pm$ 0.73} & 84.87{\scriptsize $\pm$ 1.60} & 90.49{\scriptsize $\pm$ 0.04} & 83.90{\scriptsize $\pm$ 0.28} & 84.56{\scriptsize $\pm$ 0.24} & \best{93.46{\scriptsize $\pm$ 0.14}} & 87.87{\scriptsize $\pm$ 0.47} & 85.82{\scriptsize $\pm$ 0.86} & 74.84{\scriptsize $\pm$ 1.19} & \secondb{92.55{\scriptsize $\pm$ 0.25}} \\
 & UCI & 90.35{\scriptsize $\pm$ 0.50} & 64.90{\scriptsize $\pm$ 9.03} & 82.54{\scriptsize $\pm$ 0.41} & 94.05{\scriptsize $\pm$ 0.37} & 93.93{\scriptsize $\pm$ 0.04} & 89.71{\scriptsize $\pm$ 2.88} & 91.98{\scriptsize $\pm$ 0.43} & \best{94.59{\scriptsize $\pm$ 0.22}} & 86.18{\scriptsize $\pm$ 0.30} & 88.04{\scriptsize $\pm$ 0.08} & 88.31{\scriptsize $\pm$ 0.17} & \secondb{94.15{\scriptsize $\pm$ 0.17}} \\
 & Flights & 91.92{\scriptsize $\pm$ 1.17} & 92.01{\scriptsize $\pm$ 1.02} & 94.50{\scriptsize $\pm$ 0.07} & 94.34{\scriptsize $\pm$ 0.73} & \secondb{98.44{\scriptsize $\pm$ 0.03}} & 91.31{\scriptsize $\pm$ 0.04} & 91.21{\scriptsize $\pm$ 0.01} & \best{98.95{\scriptsize $\pm$ 0.01}} & 94.66{\scriptsize $\pm$ 0.12} & 93.89{\scriptsize $\pm$ 0.06} & 91.81{\scriptsize $\pm$ 0.16} & 98.03{\scriptsize $\pm$ 0.15} \\
 & Can.\ Parl. & 80.76{\scriptsize $\pm$ 1.34} & 71.50{\scriptsize $\pm$ 0.68} & 75.48{\scriptsize $\pm$ 1.40} & 80.12{\scriptsize $\pm$ 2.27} & 75.24{\scriptsize $\pm$ 2.82} & 74.99{\scriptsize $\pm$ 1.52} & \secondb{83.83{\scriptsize $\pm$ 0.39}} & \best{97.24{\scriptsize $\pm$ 0.33}} & 77.81{\scriptsize $\pm$ 0.96} & 82.96{\scriptsize $\pm$ 0.99} & 82.13{\scriptsize $\pm$ 0.78} & 82.88{\scriptsize $\pm$ 0.76} \\
 & US Legis. & \best{84.68{\scriptsize $\pm$ 0.17}} & 78.09{\scriptsize $\pm$ 2.08} & 74.89{\scriptsize $\pm$ 0.34} & 83.15{\scriptsize $\pm$ 0.54} & 77.54{\scriptsize $\pm$ 0.08} & 75.90{\scriptsize $\pm$ 0.51} & 76.15{\scriptsize $\pm$ 0.42} & 76.95{\scriptsize $\pm$ 0.32} & 68.83{\scriptsize $\pm$ 2.94} & 82.34{\scriptsize $\pm$ 0.12} & 82.24{\scriptsize $\pm$ 0.13} & \secondb{83.53{\scriptsize $\pm$ 0.14}} \\
 & UN Trade & 69.53{\scriptsize $\pm$ 0.49} & 66.55{\scriptsize $\pm$ 0.30} & 66.85{\scriptsize $\pm$ 0.02} & 69.47{\scriptsize $\pm$ 0.77} & 69.05{\scriptsize $\pm$ 0.05} & 64.87{\scriptsize $\pm$ 0.03} & 65.02{\scriptsize $\pm$ 1.30} & 62.34{\scriptsize $\pm$ 6.07} & 69.73{\scriptsize $\pm$ 0.10} & 77.28{\scriptsize $\pm$ 0.30} & \secondb{78.01{\scriptsize $\pm$ 0.20}} & \best{80.65{\scriptsize $\pm$ 0.68}} \\
 & UN Vote & \secondb{69.66{\scriptsize $\pm$ 0.73}} & 68.88{\scriptsize $\pm$ 0.67} & 53.63{\scriptsize $\pm$ 0.31} & \best{71.99{\scriptsize $\pm$ 1.30}} & 53.41{\scriptsize $\pm$ 0.17} & 51.41{\scriptsize $\pm$ 0.27} & 52.39{\scriptsize $\pm$ 0.15} & 59.38{\scriptsize $\pm$ 0.35} & 50.00{\scriptsize $\pm$ 0.00} & 51.87{\scriptsize $\pm$ 0.98} & 53.98{\scriptsize $\pm$ 1.00} & 66.95{\scriptsize $\pm$ 0.26} \\
 & Contact & 96.65{\scriptsize $\pm$ 0.58} & 96.44{\scriptsize $\pm$ 0.16} & 97.19{\scriptsize $\pm$ 0.03} & 96.68{\scriptsize $\pm$ 0.65} & 90.38{\scriptsize $\pm$ 0.12} & 95.52{\scriptsize $\pm$ 0.15} & 93.84{\scriptsize $\pm$ 0.06} & \best{98.57{\scriptsize $\pm$ 0.01}} & 72.65{\scriptsize $\pm$ 0.21} & 75.08{\scriptsize $\pm$ 0.52} & 71.30{\scriptsize $\pm$ 0.45} & \secondb{98.21{\scriptsize $\pm$ 0.04}} \\
\cmidrule(l){2-14}
\rowcolor{RankGray}
 & Avg.\ Rank & 5.33 & 7.83 & 7.50 & 5.25 & 5.50 & 8.42 & 7.00 & \secondb{3.08} & 8.92 & 7.75 & 8.75 & \best{2.67} \\
\midrule
\multirow{13}{*}{\emph{hist}} & Wikipedia & 81.62{\scriptsize $\pm$ 0.51} & 77.80{\scriptsize $\pm$ 0.54} & 69.91{\scriptsize $\pm$ 0.42} & 82.98{\scriptsize $\pm$ 0.24} & 67.18{\scriptsize $\pm$ 3.87} & \secondb{85.93{\scriptsize $\pm$ 1.29}} & \best{87.85{\scriptsize $\pm$ 0.42}} & 80.37{\scriptsize $\pm$ 1.55} & 73.86{\scriptsize $\pm$ 0.31} & 71.65{\scriptsize $\pm$ 0.84} & 70.38{\scriptsize $\pm$ 1.04} & 79.35{\scriptsize $\pm$ 0.80} \\
 & Reddit & 79.94{\scriptsize $\pm$ 0.52} & 78.80{\scriptsize $\pm$ 1.64} & 77.71{\scriptsize $\pm$ 0.06} & 78.48{\scriptsize $\pm$ 1.05} & 80.35{\scriptsize $\pm$ 0.10} & 80.57{\scriptsize $\pm$ 0.23} & 78.35{\scriptsize $\pm$ 0.12} & \secondb{81.15{\scriptsize $\pm$ 0.27}} & 76.99{\scriptsize $\pm$ 0.13} & 78.06{\scriptsize $\pm$ 0.07} & 75.77{\scriptsize $\pm$ 0.33} & \best{81.82{\scriptsize $\pm$ 0.33}} \\
 & MOOC & 81.34{\scriptsize $\pm$ 2.48} & 72.45{\scriptsize $\pm$ 4.05} & 81.97{\scriptsize $\pm$ 0.36} & \best{90.48{\scriptsize $\pm$ 1.77}} & 72.78{\scriptsize $\pm$ 0.62} & 77.44{\scriptsize $\pm$ 0.95} & 84.68{\scriptsize $\pm$ 1.47} & \secondb{89.12{\scriptsize $\pm$ 0.88}} & 60.42{\scriptsize $\pm$ 3.86} & 73.13{\scriptsize $\pm$ 1.49} & 70.77{\scriptsize $\pm$ 1.83} & 82.50{\scriptsize $\pm$ 1.82} \\
 & LastFM & 75.93{\scriptsize $\pm$ 2.22} & 68.70{\scriptsize $\pm$ 3.12} & 62.99{\scriptsize $\pm$ 0.43} & 54.92{\scriptsize $\pm$ 9.88} & 67.69{\scriptsize $\pm$ 0.12} & 61.47{\scriptsize $\pm$ 3.00} & 64.21{\scriptsize $\pm$ 0.35} & \best{78.54{\scriptsize $\pm$ 0.20}} & 61.08{\scriptsize $\pm$ 1.52} & 60.20{\scriptsize $\pm$ 1.50} & 61.88{\scriptsize $\pm$ 0.97} & \secondb{77.36{\scriptsize $\pm$ 0.73}} \\
 & Enron & 74.41{\scriptsize $\pm$ 1.61} & 76.64{\scriptsize $\pm$ 0.71} & 65.70{\scriptsize $\pm$ 0.37} & 74.75{\scriptsize $\pm$ 2.46} & 66.52{\scriptsize $\pm$ 0.14} & 72.16{\scriptsize $\pm$ 0.61} & \best{77.50{\scriptsize $\pm$ 0.59}} & \secondb{77.47{\scriptsize $\pm$ 0.36}} & 73.49{\scriptsize $\pm$ 1.32} & 69.98{\scriptsize $\pm$ 1.64} & 62.82{\scriptsize $\pm$ 3.52} & 74.99{\scriptsize $\pm$ 1.93} \\
 & UCI & \best{80.48{\scriptsize $\pm$ 2.67}} & 55.49{\scriptsize $\pm$ 8.21} & 61.00{\scriptsize $\pm$ 0.63} & \secondb{79.10{\scriptsize $\pm$ 1.36}} & 57.84{\scriptsize $\pm$ 0.40} & 74.64{\scriptsize $\pm$ 7.05} & 78.21{\scriptsize $\pm$ 0.53} & 76.18{\scriptsize $\pm$ 0.55} & 72.87{\scriptsize $\pm$ 0.38} & 75.01{\scriptsize $\pm$ 0.69} & 77.04{\scriptsize $\pm$ 0.59} & 67.61{\scriptsize $\pm$ 0.82} \\
 & Flights & 67.44{\scriptsize $\pm$ 0.80} & 63.01{\scriptsize $\pm$ 1.89} & 71.96{\scriptsize $\pm$ 0.18} & 68.73{\scriptsize $\pm$ 1.48} & 66.38{\scriptsize $\pm$ 0.89} & 70.40{\scriptsize $\pm$ 0.22} & 70.51{\scriptsize $\pm$ 0.13} & 67.40{\scriptsize $\pm$ 1.12} & \secondb{72.82{\scriptsize $\pm$ 0.08}} & \best{73.05{\scriptsize $\pm$ 0.24}} & 70.60{\scriptsize $\pm$ 0.48} & 62.34{\scriptsize $\pm$ 2.16} \\
 & Can.\ Parl. & 77.22{\scriptsize $\pm$ 10.36} & 69.29{\scriptsize $\pm$ 1.29} & 70.46{\scriptsize $\pm$ 2.72} & 76.92{\scriptsize $\pm$ 4.29} & 70.64{\scriptsize $\pm$ 2.13} & 71.49{\scriptsize $\pm$ 1.87} & 81.58{\scriptsize $\pm$ 0.96} & \best{97.75{\scriptsize $\pm$ 0.66}} & 84.90{\scriptsize $\pm$ 1.53} & \secondb{89.30{\scriptsize $\pm$ 1.49}} & 87.71{\scriptsize $\pm$ 0.62} & 78.19{\scriptsize $\pm$ 1.97} \\
 & US Legis. & 79.40{\scriptsize $\pm$ 4.47} & 84.25{\scriptsize $\pm$ 2.48} & 64.27{\scriptsize $\pm$ 3.26} & 78.80{\scriptsize $\pm$ 4.12} & 78.22{\scriptsize $\pm$ 4.57} & 78.27{\scriptsize $\pm$ 5.61} & 73.38{\scriptsize $\pm$ 8.17} & 77.50{\scriptsize $\pm$ 6.41} & 68.79{\scriptsize $\pm$ 5.21} & \secondb{88.85{\scriptsize $\pm$ 4.53}} & \best{92.47{\scriptsize $\pm$ 0.93}} & 87.56{\scriptsize $\pm$ 1.65} \\
 & UN Trade & 63.84{\scriptsize $\pm$ 0.87} & 62.48{\scriptsize $\pm$ 0.44} & 63.62{\scriptsize $\pm$ 0.39} & 64.69{\scriptsize $\pm$ 1.41} & 66.19{\scriptsize $\pm$ 0.37} & 61.02{\scriptsize $\pm$ 0.39} & 64.01{\scriptsize $\pm$ 1.37} & 62.09{\scriptsize $\pm$ 5.46} & 71.35{\scriptsize $\pm$ 1.19} & 79.72{\scriptsize $\pm$ 0.97} & \secondb{79.72{\scriptsize $\pm$ 1.43}} & \best{81.23{\scriptsize $\pm$ 1.62}} \\
 & UN Vote & \secondb{78.79{\scriptsize $\pm$ 0.92}} & 75.89{\scriptsize $\pm$ 1.22} & 58.80{\scriptsize $\pm$ 1.20} & 75.91{\scriptsize $\pm$ 1.64} & 51.46{\scriptsize $\pm$ 0.68} & 51.29{\scriptsize $\pm$ 1.16} & 53.65{\scriptsize $\pm$ 1.44} & 67.62{\scriptsize $\pm$ 0.64} & 50.00{\scriptsize $\pm$ 0.00} & 55.86{\scriptsize $\pm$ 2.62} & 58.93{\scriptsize $\pm$ 1.11} & \best{84.05{\scriptsize $\pm$ 2.79}} \\
 & Contact & \secondb{96.42{\scriptsize $\pm$ 0.43}} & 95.87{\scriptsize $\pm$ 0.22} & 95.76{\scriptsize $\pm$ 0.19} & 94.81{\scriptsize $\pm$ 1.47} & 83.73{\scriptsize $\pm$ 0.33} & 94.70{\scriptsize $\pm$ 0.27} & 93.25{\scriptsize $\pm$ 0.29} & \best{97.15{\scriptsize $\pm$ 0.10}} & 64.50{\scriptsize $\pm$ 0.69} & 68.73{\scriptsize $\pm$ 0.46} & 64.65{\scriptsize $\pm$ 1.00} & 95.98{\scriptsize $\pm$ 0.15} \\
\cmidrule(l){2-14}
\rowcolor{RankGray}
 & Avg.\ Rank & 4.75 & 7.25 & 8.42 & 5.50 & 8.58 & 7.33 & 5.50 & \secondb{4.42} & 8.42 & 6.50 & 7.00 & \best{4.33} \\
\midrule
\multirow{13}{*}{\emph{ind}} & Wikipedia & 72.75{\scriptsize $\pm$ 0.63} & 70.45{\scriptsize $\pm$ 0.77} & 72.35{\scriptsize $\pm$ 0.45} & \secondb{81.45{\scriptsize $\pm$ 0.33}} & 69.43{\scriptsize $\pm$ 4.24} & 74.14{\scriptsize $\pm$ 3.26} & \best{84.76{\scriptsize $\pm$ 0.38}} & 78.68{\scriptsize $\pm$ 2.75} & 57.84{\scriptsize $\pm$ 0.47} & 56.98{\scriptsize $\pm$ 0.45} & 56.61{\scriptsize $\pm$ 0.44} & 77.62{\scriptsize $\pm$ 2.29} \\
 & Reddit & 83.23{\scriptsize $\pm$ 0.29} & 82.23{\scriptsize $\pm$ 1.46} & 85.55{\scriptsize $\pm$ 0.07} & 82.19{\scriptsize $\pm$ 1.54} & \best{88.32{\scriptsize $\pm$ 0.12}} & 73.02{\scriptsize $\pm$ 1.12} & 85.06{\scriptsize $\pm$ 0.08} & 85.68{\scriptsize $\pm$ 0.41} & 74.38{\scriptsize $\pm$ 0.51} & 78.23{\scriptsize $\pm$ 0.33} & 78.12{\scriptsize $\pm$ 0.20} & \secondb{87.60{\scriptsize $\pm$ 0.30}} \\
 & MOOC & 67.88{\scriptsize $\pm$ 1.79} & 59.54{\scriptsize $\pm$ 4.35} & 75.56{\scriptsize $\pm$ 0.70} & \best{83.13{\scriptsize $\pm$ 3.07}} & 71.58{\scriptsize $\pm$ 0.64} & 74.63{\scriptsize $\pm$ 0.79} & 77.51{\scriptsize $\pm$ 1.11} & \secondb{82.55{\scriptsize $\pm$ 0.68}} & 51.71{\scriptsize $\pm$ 2.49} & 56.38{\scriptsize $\pm$ 0.66} & 57.15{\scriptsize $\pm$ 0.62} & 74.66{\scriptsize $\pm$ 3.07} \\
 & LastFM & 63.34{\scriptsize $\pm$ 1.75} & 61.19{\scriptsize $\pm$ 1.55} & 62.99{\scriptsize $\pm$ 0.41} & 53.18{\scriptsize $\pm$ 4.97} & 67.77{\scriptsize $\pm$ 0.15} & 55.40{\scriptsize $\pm$ 2.81} & 60.77{\scriptsize $\pm$ 0.29} & \secondb{68.50{\scriptsize $\pm$ 0.59}} & 52.21{\scriptsize $\pm$ 0.79} & 52.23{\scriptsize $\pm$ 0.78} & 51.77{\scriptsize $\pm$ 0.31} & \best{69.65{\scriptsize $\pm$ 2.74}} \\
 & Enron & 72.78{\scriptsize $\pm$ 1.45} & 72.65{\scriptsize $\pm$ 1.36} & 64.91{\scriptsize $\pm$ 0.42} & 71.51{\scriptsize $\pm$ 3.26} & \secondb{76.72{\scriptsize $\pm$ 0.28}} & 70.77{\scriptsize $\pm$ 0.70} & 72.83{\scriptsize $\pm$ 0.49} & 75.45{\scriptsize $\pm$ 0.71} & 66.21{\scriptsize $\pm$ 1.20} & 67.32{\scriptsize $\pm$ 1.92} & 58.12{\scriptsize $\pm$ 3.21} & \best{79.55{\scriptsize $\pm$ 0.71}} \\
 & UCI & 63.20{\scriptsize $\pm$ 1.39} & 52.22{\scriptsize $\pm$ 1.92} & 60.51{\scriptsize $\pm$ 0.43} & 64.22{\scriptsize $\pm$ 0.94} & 57.67{\scriptsize $\pm$ 0.43} & \secondb{70.11{\scriptsize $\pm$ 3.98}} & \best{74.34{\scriptsize $\pm$ 0.58}} & 65.51{\scriptsize $\pm$ 0.74} & 50.54{\scriptsize $\pm$ 0.35} & 54.48{\scriptsize $\pm$ 0.39} & 56.16{\scriptsize $\pm$ 0.08} & 57.52{\scriptsize $\pm$ 0.44} \\
 & Flights & 68.28{\scriptsize $\pm$ 1.44} & 64.04{\scriptsize $\pm$ 2.14} & 72.87{\scriptsize $\pm$ 0.28} & 71.91{\scriptsize $\pm$ 1.68} & 70.12{\scriptsize $\pm$ 0.89} & 72.48{\scriptsize $\pm$ 0.20} & 72.36{\scriptsize $\pm$ 0.19} & 68.36{\scriptsize $\pm$ 1.58} & \secondb{73.15{\scriptsize $\pm$ 0.30}} & \best{75.33{\scriptsize $\pm$ 0.54}} & 72.68{\scriptsize $\pm$ 0.46} & 66.71{\scriptsize $\pm$ 1.01} \\
 & Can.\ Parl. & 67.81{\scriptsize $\pm$ 12.02} & 63.83{\scriptsize $\pm$ 0.82} & 69.54{\scriptsize $\pm$ 1.78} & 74.82{\scriptsize $\pm$ 4.48} & 73.30{\scriptsize $\pm$ 1.08} & 72.13{\scriptsize $\pm$ 0.80} & 72.14{\scriptsize $\pm$ 0.98} & \best{97.53{\scriptsize $\pm$ 0.44}} & 77.81{\scriptsize $\pm$ 0.75} & \secondb{83.69{\scriptsize $\pm$ 1.26}} & 82.15{\scriptsize $\pm$ 0.38} & 76.29{\scriptsize $\pm$ 3.18} \\
 & US Legis. & 81.63{\scriptsize $\pm$ 3.74} & 83.66{\scriptsize $\pm$ 2.33} & 62.35{\scriptsize $\pm$ 3.80} & 73.73{\scriptsize $\pm$ 4.72} & 74.50{\scriptsize $\pm$ 4.20} & 76.55{\scriptsize $\pm$ 8.69} & 71.15{\scriptsize $\pm$ 9.06} & 75.45{\scriptsize $\pm$ 5.69} & 71.18{\scriptsize $\pm$ 3.09} & \secondb{89.21{\scriptsize $\pm$ 1.90}} & \best{92.17{\scriptsize $\pm$ 0.19}} & 88.67{\scriptsize $\pm$ 1.06} \\
 & UN Trade & 65.83{\scriptsize $\pm$ 0.96} & 64.43{\scriptsize $\pm$ 0.20} & 70.71{\scriptsize $\pm$ 0.31} & 66.95{\scriptsize $\pm$ 1.90} & 73.89{\scriptsize $\pm$ 0.25} & 67.14{\scriptsize $\pm$ 0.41} & 67.93{\scriptsize $\pm$ 1.41} & 62.85{\scriptsize $\pm$ 5.23} & 71.25{\scriptsize $\pm$ 0.64} & 79.40{\scriptsize $\pm$ 0.62} & \secondb{80.30{\scriptsize $\pm$ 0.58}} & \best{84.53{\scriptsize $\pm$ 0.74}} \\
 & UN Vote & \secondb{75.70{\scriptsize $\pm$ 1.40}} & 74.74{\scriptsize $\pm$ 1.21} & 55.59{\scriptsize $\pm$ 1.03} & 74.87{\scriptsize $\pm$ 1.56} & 54.20{\scriptsize $\pm$ 1.68} & 53.29{\scriptsize $\pm$ 1.28} & 51.44{\scriptsize $\pm$ 0.64} & 57.19{\scriptsize $\pm$ 0.29} & 50.00{\scriptsize $\pm$ 0.00} & 51.19{\scriptsize $\pm$ 2.24} & 57.90{\scriptsize $\pm$ 2.11} & \best{84.54{\scriptsize $\pm$ 0.71}} \\
 & Contact & 94.57{\scriptsize $\pm$ 0.73} & 94.65{\scriptsize $\pm$ 0.30} & 94.81{\scriptsize $\pm$ 0.22} & 92.87{\scriptsize $\pm$ 1.87} & 88.95{\scriptsize $\pm$ 0.28} & 92.69{\scriptsize $\pm$ 0.32} & 91.29{\scriptsize $\pm$ 0.30} & \secondb{95.30{\scriptsize $\pm$ 0.35}} & 55.37{\scriptsize $\pm$ 0.55} & 59.52{\scriptsize $\pm$ 0.70} & 53.94{\scriptsize $\pm$ 0.57} & \best{95.79{\scriptsize $\pm$ 0.40}} \\
\cmidrule(l){2-14}
\rowcolor{RankGray}
 & Avg.\ Rank & 6.42 & 7.83 & 6.50 & 5.92 & 6.08 & 7.08 & 5.92 & \secondb{4.42} & 9.17 & 7.42 & 7.67 & \best{3.58} \\
\bottomrule
\end{tabular}%
}
\end{table*}
\section{Experiments}
\noindent\textbf{Datasets and Baselines.} We evaluate on 12 dynamic graph datasets from the DyGLib benchmark~\citep{yu2023towards}: Wikipedia, Reddit, MOOC, LastFM, Enron, UCI, Can.~Parl., US~Legis., UN~Trade, UN~Vote, Contact and Flights, each split chronologically into 70\%/15\%/15\% for training, validation and testing. Details are in Appendix~\ref{app:datasets}. We compare against eleven baselines in two families. The first consists of eight supervised temporal graph learning methods: three memory-based methods, JODIE~\citep{kumar2019predicting}, DyRep~\citep{trivedi2019dyrep} and TGN~\citep{rossi2020temporal}; a random-walk-based method, CAWN~\citep{wang2021inductive}; a GNN-based method, TGAT~\citep{xu2020inductive}; an MLP-based method, GraphMixer~\citep{cong2023we}; and two Transformer-based methods, TCL~\citep{wang2021tcl} and DyGFormer~\citep{yu2023towards}. The second consists of three self-supervised dynamic graph representation learning methods: CLDG~\citep{xu2023cldg}, DVGMAE~\citep{gao2025dvgmae} and MaskDGNN~\citep{he2025maskdgnn}. For dynamic node classification, we additionally include two discrete-time DGNNs, EvolveGCN-H~\citep{pareja2020evolvegcn} and ROLAND~\citep{you2022roland}, and compare against SG-JEPA~\citep{zhang2026scalable}, the most closely related joint-embedding predictive architecture for dynamic graphs.
\begin{table*}[h]
\centering
\caption{\textbf{Inductive} AUC-ROC for dynamic link prediction in the inductive (new-node) setting with random negative edge sampling. \best{Bold red} and \secondb{underlined blue} indicate the best and second-best results, respectively.}
\label{tab:auc_inductive}
\resizebox{\textwidth}{!}{%
\begin{tabular}{lccccccccccc>{\columncolor{OursTint}}c}
\toprule
Datasets & JODIE & DyRep & TGAT & TGN & CAWN & TCL & GraphMixer & DyGFormer & CLDG & MaskDGNN & DVGMAE & \method \\
\midrule

Wikipedia
& 94.06{\scriptsize $\pm$ 0.23}
& 91.46{\scriptsize $\pm$ 0.24}
& 78.29{\scriptsize $\pm$ 0.67}
& 97.22{\scriptsize $\pm$ 0.07}
& 96.93{\scriptsize $\pm$ 0.02}
& 96.74{\scriptsize $\pm$ 0.09}
& 95.89{\scriptsize $\pm$ 0.06}
& \secondb{97.42{\scriptsize $\pm$ 0.02}}
& 86.63{\scriptsize $\pm$ 0.42}
& 84.35{\scriptsize $\pm$ 0.20}
& 80.43{\scriptsize $\pm$ 1.29}
& \best{97.52{\scriptsize $\pm$ 0.03}} \\

Reddit
& 96.01{\scriptsize $\pm$ 0.10}
& 95.25{\scriptsize $\pm$ 0.29}
& 67.96{\scriptsize $\pm$ 1.62}
& 96.12{\scriptsize $\pm$ 0.61}
& 96.48{\scriptsize $\pm$ 0.03}
& 95.95{\scriptsize $\pm$ 0.09}
& 95.36{\scriptsize $\pm$ 0.11}
& \secondb{96.56{\scriptsize $\pm$ 0.02}}
& 94.33{\scriptsize $\pm$ 0.08}
& 94.66{\scriptsize $\pm$ 0.06}
& 93.26{\scriptsize $\pm$ 0.21}
& \best{97.44{\scriptsize $\pm$ 0.10}} \\

MOOC
& 81.86{\scriptsize $\pm$ 0.78}
& 79.41{\scriptsize $\pm$ 2.36}
& 80.43{\scriptsize $\pm$ 0.30}
& \best{91.44{\scriptsize $\pm$ 0.49}}
& 82.43{\scriptsize $\pm$ 0.28}
& 82.33{\scriptsize $\pm$ 0.65}
& 83.09{\scriptsize $\pm$ 0.18}
& \secondb{87.92{\scriptsize $\pm$ 0.29}}
& 63.33{\scriptsize $\pm$ 0.58}
& 72.03{\scriptsize $\pm$ 0.39}
& 66.51{\scriptsize $\pm$ 0.95}
& 79.63{\scriptsize $\pm$ 0.44} \\

LastFM
& 81.89{\scriptsize $\pm$ 0.69}
& 81.36{\scriptsize $\pm$ 1.62}
& 76.83{\scriptsize $\pm$ 0.14}
& 62.44{\scriptsize $\pm$ 5.05}
& 87.88{\scriptsize $\pm$ 0.05}
& 76.36{\scriptsize $\pm$ 1.04}
& 80.10{\scriptsize $\pm$ 0.16}
& \secondb{93.05{\scriptsize $\pm$ 0.05}}
& 67.08{\scriptsize $\pm$ 1.27}
& 66.47{\scriptsize $\pm$ 1.02}
& 63.90{\scriptsize $\pm$ 0.57}
& \best{93.50{\scriptsize $\pm$ 0.22}} \\

Enron
& 82.81{\scriptsize $\pm$ 1.85}
& 74.13{\scriptsize $\pm$ 2.31}
& 69.84{\scriptsize $\pm$ 0.47}
& 74.17{\scriptsize $\pm$ 2.79}
& \secondb{87.60{\scriptsize $\pm$ 0.25}}
& 81.23{\scriptsize $\pm$ 0.33}
& 77.88{\scriptsize $\pm$ 0.15}
& \best{90.73{\scriptsize $\pm$ 0.10}}
& 80.10{\scriptsize $\pm$ 1.15}
& 76.05{\scriptsize $\pm$ 1.65}
& 62.54{\scriptsize $\pm$ 0.60}
& 87.45{\scriptsize $\pm$ 0.26} \\

UCI
& 77.47{\scriptsize $\pm$ 1.78}
& 58.26{\scriptsize $\pm$ 6.40}
& 77.93{\scriptsize $\pm$ 0.49}
& 86.21{\scriptsize $\pm$ 2.79}
& 89.65{\scriptsize $\pm$ 0.16}
& 79.89{\scriptsize $\pm$ 0.98}
& 88.84{\scriptsize $\pm$ 0.15}
& \secondb{92.03{\scriptsize $\pm$ 0.09}}
& 75.80{\scriptsize $\pm$ 0.37}
& 76.76{\scriptsize $\pm$ 0.18}
& 77.16{\scriptsize $\pm$ 0.11}
& \best{92.32{\scriptsize $\pm$ 0.13}} \\

Flights
& 90.02{\scriptsize $\pm$ 1.59}
& 86.79{\scriptsize $\pm$ 1.09}
& 87.50{\scriptsize $\pm$ 0.11}
& 87.28{\scriptsize $\pm$ 0.79}
& \secondb{97.20{\scriptsize $\pm$ 0.03}}
& 84.18{\scriptsize $\pm$ 0.05}
& 83.85{\scriptsize $\pm$ 0.12}
& \best{98.06{\scriptsize $\pm$ 0.02}}
& 90.27{\scriptsize $\pm$ 0.28}
& 89.06{\scriptsize $\pm$ 0.08}
& 84.55{\scriptsize $\pm$ 0.44}
& 93.98{\scriptsize $\pm$ 0.46} \\

Can.\ Parl.
& 54.34{\scriptsize $\pm$ 2.88}
& 56.87{\scriptsize $\pm$ 1.73}
& 59.57{\scriptsize $\pm$ 1.46}
& 61.61{\scriptsize $\pm$ 1.38}
& 63.27{\scriptsize $\pm$ 0.54}
& 61.04{\scriptsize $\pm$ 0.27}
& 60.06{\scriptsize $\pm$ 0.70}
& \best{96.51{\scriptsize $\pm$ 0.55}}
& 61.78{\scriptsize $\pm$ 0.68}
& 60.35{\scriptsize $\pm$ 1.78}
& \secondb{63.45{\scriptsize $\pm$ 0.58}}
& 62.63{\scriptsize $\pm$ 1.08} \\

US Legis.
& \secondb{66.20{\scriptsize $\pm$ 0.44}}
& 60.56{\scriptsize $\pm$ 0.98}
& 47.87{\scriptsize $\pm$ 1.00}
& 63.91{\scriptsize $\pm$ 0.42}
& 54.53{\scriptsize $\pm$ 1.45}
& 51.35{\scriptsize $\pm$ 0.55}
& 51.97{\scriptsize $\pm$ 1.39}
& 54.20{\scriptsize $\pm$ 0.41}
& 59.01{\scriptsize $\pm$ 4.64}
& 65.25{\scriptsize $\pm$ 0.28}
& 65.29{\scriptsize $\pm$ 0.30}
& \best{66.75{\scriptsize $\pm$ 1.29}} \\

UN Trade
& 61.11{\scriptsize $\pm$ 0.34}
& 58.65{\scriptsize $\pm$ 1.10}
& 56.94{\scriptsize $\pm$ 1.45}
& 61.87{\scriptsize $\pm$ 0.65}
& 67.06{\scriptsize $\pm$ 0.20}
& 63.59{\scriptsize $\pm$ 0.06}
& 62.96{\scriptsize $\pm$ 0.85}
& 53.59{\scriptsize $\pm$ 3.14}
& 67.45{\scriptsize $\pm$ 0.21}
& \secondb{74.92{\scriptsize $\pm$ 0.37}}
& \best{76.31{\scriptsize $\pm$ 0.31}}
& 67.42{\scriptsize $\pm$ 1.82} \\

UN Vote
& \secondb{58.19{\scriptsize $\pm$ 1.79}}
& 56.75{\scriptsize $\pm$ 0.91}
& 51.31{\scriptsize $\pm$ 0.51}
& \best{61.08{\scriptsize $\pm$ 0.50}}
& 49.23{\scriptsize $\pm$ 0.63}
& 48.95{\scriptsize $\pm$ 0.33}
& 50.77{\scriptsize $\pm$ 0.25}
& 56.37{\scriptsize $\pm$ 0.25}
& 49.86{\scriptsize $\pm$ 0.27}
& 48.42{\scriptsize $\pm$ 0.46}
& 49.75{\scriptsize $\pm$ 0.75}
& 54.84{\scriptsize $\pm$ 1.17} \\

Contact
& 95.87{\scriptsize $\pm$ 0.65}
& 92.36{\scriptsize $\pm$ 0.38}
& 96.40{\scriptsize $\pm$ 0.05}
& 94.39{\scriptsize $\pm$ 0.85}
& 89.51{\scriptsize $\pm$ 0.27}
& 94.73{\scriptsize $\pm$ 0.13}
& 92.63{\scriptsize $\pm$ 0.05}
& \best{98.30{\scriptsize $\pm$ 0.01}}
& 69.24{\scriptsize $\pm$ 0.89}
& 70.23{\scriptsize $\pm$ 0.54}
& 65.79{\scriptsize $\pm$ 0.61}
& \secondb{96.75{\scriptsize $\pm$ 0.11}} \\

\cmidrule(l){1-13}
\rowcolor{RankGray}
Avg.\ Rank
& 5.67
& 8.25
& 8.75
& 5.67
& 4.67
& 7.17
& 7.08
& \secondb{3.25}
& 7.92
& 8.25
& 8.50
& \best{2.83} \\

\bottomrule
\end{tabular}%
}
\vspace{-15pt}
\end{table*}

\textbf{Implementation Details.} For a fair comparison, we reproduce all baselines in a unified pipeline built on DyGLib \citep{yu2023towards}, with the same event-bin discretization, the same chronological split and the same negative sampling protocol. We set the same history window for \method~ and the snapshot-based baselines to fairly compare their dynamic graph modeling ability. We train each model for up to 100 epochs
with early stopping and select the best-performing checkpoint for testing. We repeat each experiment 5 times with different random seeds and report the mean and standard deviation. Details in Appendix~\ref{app:impl}.

\textbf{Evaluation Details.} We evaluate dynamic link prediction in both transductive and inductive settings. The transductive setting considers future links between nodes observed during training, while the inductive setting evaluates links involving unseen nodes. Following DyGLib~\citep{yu2023towards}, we use random (\textit{rnd}), historical (\textit{hist}), and inductive (\textit{ind}) negative sampling, and report Average Precision (AP) and AUC-ROC. For dynamic node classification, we use labelled-node training ratios of 40\%, 60\%, and 80\%, and report Macro-F1 and Micro-F1 on the held-out test nodes. More details are given in Appendix~\ref{app:evaluation-settings}.
\begin{table*}[h]
\vspace{-10pt}
\centering
\caption{\textbf{Node classification on dynamic graphs.} Macro-F1 and Micro-F1 (\%) on the held-out test nodes, reported as mean $\pm$ std over 5 seeds at labelled-node training ratios of 40\%, 60\%, and 80\%. \best{Bold red} and \secondb{underlined blue} indicate the best and second-best results, respectively.}
\label{tab:node_classification}
\resizebox{\textwidth}{!}{%
\begin{tabular}{lllccccccccccc>{\columncolor{OursTint}}c}
\toprule
Metric & Dataset & Ratio & TGAT& CAWN & TCL & GraphMixer & DyGFormer & EvolveGCN-H & ROLAND & CLDG & MaskDGNN & DVGMAE & SG-JEPA & \method \\
\midrule
\multirow{6}{*}{Macro-F1}
& \multirow{3}{*}{DBLP}
& 40\% & 68.54{\scriptsize $\pm$ 0.60} & 64.71{\scriptsize $\pm$ 0.92} & 69.76{\scriptsize $\pm$ 1.09} & 71.69{\scriptsize $\pm$ 0.79} & 67.88{\scriptsize $\pm$ 0.72} & 26.10{\scriptsize $\pm$ 11.19} & 47.14{\scriptsize $\pm$ 20.87} & 60.70{\scriptsize $\pm$ 1.34} & 70.98{\scriptsize $\pm$ 0.84} & 60.36{\scriptsize $\pm$ 0.83} & \secondb{73.96{\scriptsize $\pm$ 0.59}} & \best{75.45{\scriptsize $\pm$ 0.35}} \\
& & 60\% & 71.60{\scriptsize $\pm$ 0.61} & 66.32{\scriptsize $\pm$ 0.60} & 72.33{\scriptsize $\pm$ 0.68} & 73.80{\scriptsize $\pm$ 0.52} & 69.99{\scriptsize $\pm$ 0.62} & 32.22{\scriptsize $\pm$ 19.39} & 58.32{\scriptsize $\pm$ 13.49} & 59.70{\scriptsize $\pm$ 1.95} & 71.91{\scriptsize $\pm$ 0.24} & 60.36{\scriptsize $\pm$ 0.81} & \secondb{75.88{\scriptsize $\pm$ 0.36}} & \best{77.13{\scriptsize $\pm$ 0.51}} \\
& & 80\% & 73.12{\scriptsize $\pm$ 0.56} & 67.66{\scriptsize $\pm$ 1.11} & 73.59{\scriptsize $\pm$ 0.77} & 75.41{\scriptsize $\pm$ 0.77} & 71.46{\scriptsize $\pm$ 0.80} & 37.47{\scriptsize $\pm$ 9.69} & 52.91{\scriptsize $\pm$ 12.28} & 61.77{\scriptsize $\pm$ 2.20} & 73.88{\scriptsize $\pm$ 1.21} & 60.38{\scriptsize $\pm$ 1.00} & \secondb{77.55{\scriptsize $\pm$ 0.33}} & \best{78.13{\scriptsize $\pm$ 0.23}} \\
\cmidrule(lr){2-15}

& \multirow{3}{*}{Tmall}
& 40\% & 47.36{\scriptsize $\pm$ 0.78} & 45.49{\scriptsize $\pm$ 0.35} & 49.12{\scriptsize $\pm$ 0.49} & 49.20{\scriptsize $\pm$ 0.94} & 33.31{\scriptsize $\pm$ 6.17} & 35.80{\scriptsize $\pm$ 9.81} & 41.46{\scriptsize $\pm$ 8.38} & 32.06{\scriptsize $\pm$ 1.83} & 48.35{\scriptsize $\pm$ 1.60} & 36.23{\scriptsize $\pm$ 1.02} & \secondb{54.16{\scriptsize $\pm$ 0.36}} & \best{57.37{\scriptsize $\pm$ 0.37}} \\
& & 60\% & 49.05{\scriptsize $\pm$ 0.26} & 47.28{\scriptsize $\pm$ 0.56} & 50.10{\scriptsize $\pm$ 0.58} & 51.50{\scriptsize $\pm$ 0.75} & 32.65{\scriptsize $\pm$ 2.84} & 36.13{\scriptsize $\pm$ 6.82} & 40.87{\scriptsize $\pm$ 6.29} & 30.10{\scriptsize $\pm$ 1.44} & 50.83{\scriptsize $\pm$ 1.03} & 35.47{\scriptsize $\pm$ 1.19} & \secondb{55.81{\scriptsize $\pm$ 0.34}} & \best{59.17{\scriptsize $\pm$ 0.49}} \\
& & 80\% & 49.86{\scriptsize $\pm$ 0.44} & 48.23{\scriptsize $\pm$ 0.82} & 50.56{\scriptsize $\pm$ 0.78} & 52.92{\scriptsize $\pm$ 0.41} & 34.73{\scriptsize $\pm$ 2.49} & 39.41{\scriptsize $\pm$ 3.33} & 40.74{\scriptsize $\pm$ 6.82} & 31.00{\scriptsize $\pm$ 1.70} & 51.39{\scriptsize $\pm$ 1.52} & 35.40{\scriptsize $\pm$ 1.30} & \secondb{57.24{\scriptsize $\pm$ 0.39}} & \best{60.52{\scriptsize $\pm$ 0.39}} \\
\midrule

\multirow{6}{*}{Micro-F1}
& \multirow{3}{*}{DBLP}
& 40\% & 69.59{\scriptsize $\pm$ 0.48} & 66.24{\scriptsize $\pm$ 0.41} & 70.35{\scriptsize $\pm$ 1.55} & 72.15{\scriptsize $\pm$ 0.67} & 68.31{\scriptsize $\pm$ 0.69} & 43.97{\scriptsize $\pm$ 7.17} & 59.19{\scriptsize $\pm$ 14.41} & 62.97{\scriptsize $\pm$ 0.85} & 71.01{\scriptsize $\pm$ 0.67} & 62.41{\scriptsize $\pm$ 0.58} & \secondb{74.20{\scriptsize $\pm$ 0.55}} & \best{75.79{\scriptsize $\pm$ 0.24}} \\
& & 60\% & 71.72{\scriptsize $\pm$ 0.50} & 67.94{\scriptsize $\pm$ 0.56} & 72.84{\scriptsize $\pm$ 0.80} & 74.15{\scriptsize $\pm$ 0.23} & 70.25{\scriptsize $\pm$ 0.31} & 45.96{\scriptsize $\pm$ 14.30} & 66.06{\scriptsize $\pm$ 4.61} & 62.40{\scriptsize $\pm$ 1.45} & 72.07{\scriptsize $\pm$ 0.32} & 62.42{\scriptsize $\pm$ 0.58} & \secondb{76.17{\scriptsize $\pm$ 0.18}} & \best{77.52{\scriptsize $\pm$ 0.39}} \\
& & 80\% & 73.43{\scriptsize $\pm$ 0.50} & 68.67{\scriptsize $\pm$ 0.83} & 73.78{\scriptsize $\pm$ 0.88} & 75.60{\scriptsize $\pm$ 0.26} & 70.99{\scriptsize $\pm$ 0.97} & 51.36{\scriptsize $\pm$ 6.31} & 64.18{\scriptsize $\pm$ 4.32} & 63.64{\scriptsize $\pm$ 1.33} & 73.41{\scriptsize $\pm$ 1.00} & 62.46{\scriptsize $\pm$ 0.91} & \secondb{77.57{\scriptsize $\pm$ 0.21}} & \best{78.42{\scriptsize $\pm$ 0.23}} \\
\cmidrule(lr){2-15}

& \multirow{3}{*}{Tmall}
& 40\% & 53.28{\scriptsize $\pm$ 0.80} & 50.82{\scriptsize $\pm$ 0.80} & 54.56{\scriptsize $\pm$ 0.50} & 54.31{\scriptsize $\pm$ 0.66} & 45.61{\scriptsize $\pm$ 2.26} & 50.60{\scriptsize $\pm$ 7.74} & 54.41{\scriptsize $\pm$ 4.10} & 45.23{\scriptsize $\pm$ 0.66} & 55.43{\scriptsize $\pm$ 1.39} & 47.94{\scriptsize $\pm$ 0.64} & \secondb{58.18{\scriptsize $\pm$ 0.76}} & \best{61.88{\scriptsize $\pm$ 0.30}} \\
& & 60\% & 54.88{\scriptsize $\pm$ 0.41} & 53.03{\scriptsize $\pm$ 0.49} & 56.19{\scriptsize $\pm$ 0.75} & 56.69{\scriptsize $\pm$ 1.17} & 45.25{\scriptsize $\pm$ 2.39} & 52.33{\scriptsize $\pm$ 4.71} & 54.91{\scriptsize $\pm$ 3.32} & 44.34{\scriptsize $\pm$ 0.64} & 57.51{\scriptsize $\pm$ 0.48} & 47.89{\scriptsize $\pm$ 0.52} & \secondb{59.93{\scriptsize $\pm$ 0.83}} & \best{63.72{\scriptsize $\pm$ 0.43}} \\
& & 80\% & 55.97{\scriptsize $\pm$ 0.44} & 54.28{\scriptsize $\pm$ 0.61} & 57.05{\scriptsize $\pm$ 0.57} & 57.90{\scriptsize $\pm$ 0.57} & 47.22{\scriptsize $\pm$ 0.81} & 54.13{\scriptsize $\pm$ 1.68} & 54.68{\scriptsize $\pm$ 3.83} & 45.15{\scriptsize $\pm$ 0.80} & 57.68{\scriptsize $\pm$ 1.45} & 47.43{\scriptsize $\pm$ 1.35} & \secondb{61.71{\scriptsize $\pm$ 0.53}} & \best{65.08{\scriptsize $\pm$ 0.37}} \\
\bottomrule
\end{tabular}%
\vspace{-10pt}
}
\end{table*}
\begin{wrapfigure}{r}{10cm}
    \vspace{-10pt}
    \centering
    \includegraphics[
        width=\linewidth
    ]{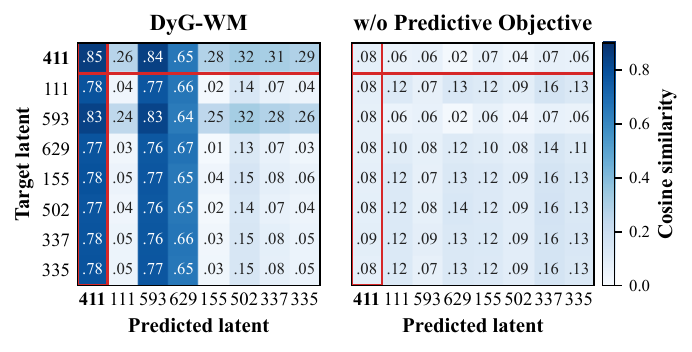}
    \vspace{-25pt}
    \caption{Latent alignment on Can.\ Parl. with and without the predictive objective.}
    \label{fig:ablation_heatmap_canparl}
    \vspace{-30pt}
\end{wrapfigure}
\subsection{Main Results}
\noindent\textbf{Overall Performance.} Tables~\ref{tab:auc_main} and~\ref{tab:auc_inductive} show that \method achieves the best average rank under all three transductive negative-sampling strategies, with ranks of 2.67, 4.33, and 3.58, and also ranks first in the new-node setting with 2.83. The advantage is more apparent under inductive negatives, where \method performs best on several datasets, including UN Trade and UN Vote. For dynamic node classification, Table~\ref{tab:node_classification} shows that \method gives the best Macro-F1 and Micro-F1 on both DBLP and Tmall across all three training ratios. Overall, the results show that \method performs well on both link prediction and node classification. More details in Appendix~\ref{additional experiment}.
\subsection{Ablation Study}
\noindent\textbf{Component Contributions.}
Table~\ref{tab:ablation} shows that removing any component lowers AUC-ROC on all three datasets. The largest drop comes from removing the path-signature encoder, from 74.99 to 59.09 on Enron, 78.19 to 60.31 on Can.\ Parl., and 77.36 to 70.17 on LastFM. Historical context and the predictive objective also provide consistent gains.
\begin{wraptable}{r}{8.5cm}
\vspace{-20pt}
\centering
\scriptsize
\caption{Results (AUC-ROC) of ablations.}
\label{tab:ablation}
\resizebox{\linewidth}{!}{%
\begin{tabular}{lccc}
\toprule
\rowcolor{gray!30}
Settings & Enron & Can.\ Parl. & LastFM \\
\midrule
w/o Historical Context
& 71.53{\scriptsize $\pm$ 3.02}
& 76.08{\scriptsize $\pm$ 2.77}
& 75.11{\scriptsize $\pm$ 0.52} \\

w/o Path Signature
& 59.09{\scriptsize $\pm$ 5.52}
& 60.31{\scriptsize $\pm$ 4.65}
& 70.17{\scriptsize $\pm$ 0.87} \\

w/o Predictive Objective
& 71.69{\scriptsize $\pm$ 3.94}
& 75.58{\scriptsize $\pm$ 2.11}
& 75.93{\scriptsize $\pm$ 1.32} \\
\midrule

\rowcolor{RankGray}
\method
& \best{74.99{\scriptsize $\pm$ 1.93}}
& \best{78.19{\scriptsize $\pm$ 1.97}}
& \best{77.36{\scriptsize $\pm$ 0.73}} \\
\bottomrule
\end{tabular}%
}
\end{wraptable}
\noindent\textbf{Predictive Alignment.}
Figure~\ref{fig:ablation_heatmap_canparl} shows that the JEPA objective substantially improves alignment between predicted and target latents. With the full model, many cosine similarities are around $0.64$-$0.85$, while removing the objective reduces most values to near zero. This matches the performance drop observed in the ablation table. More details in Appendix~\ref{additional experiment}.
\begin{wrapfigure}{r}{10.5cm}
    \vspace{-40pt}
    \centering
    \begin{minipage}[t]{0.49\linewidth}
        \centering
        \includegraphics[
            width=\linewidth
        ]{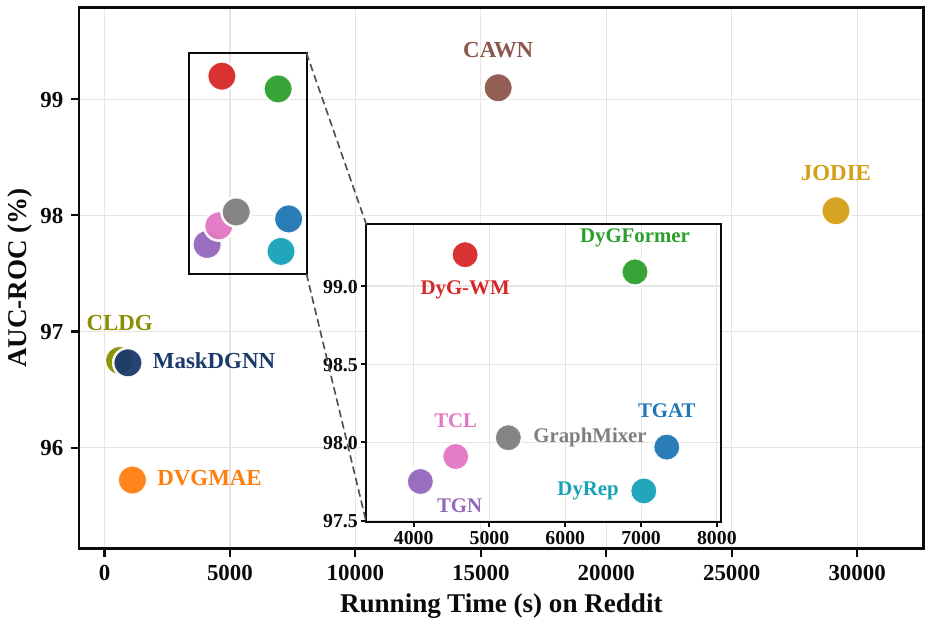}
    \end{minipage}
    \hfill
    \begin{minipage}[t]{0.49\linewidth}
        \centering
        \includegraphics[
            width=\linewidth
        ]{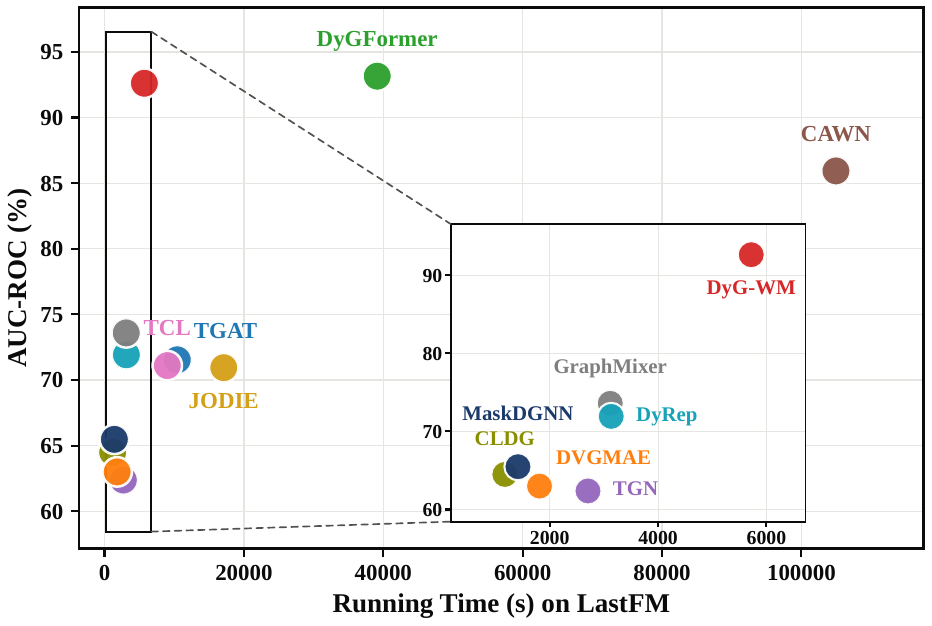}
    \end{minipage}
    \caption{AUC-ROC-efficiency trade-off on Reddit and LastFM.}
    \vspace{0pt}
    \label{fig:efficiency}
\end{wrapfigure}
\subsection{Efficiency}
As shown in Figure~\ref{fig:efficiency}, Reddit and LastFM exhibit different accuracy--efficiency trade-offs. On Reddit, several strong baselines already approach 99 AUC-ROC, so the main difference lies in the computation needed to reach this accuracy range. \method attains the highest AUC-ROC while training faster than other high-accuracy models such as DyGFormer and CAWN. On LastFM, the trade-off is more pronounced: DyGFormer remains slightly more accurate, but requires substantially more training time, whereas faster baselines incur large accuracy losses. \method stays close to the best accuracy while using a moderate training budget. Across both datasets, no compared method is simultaneously faster and more accurate than \method, placing it on the accuracy-efficiency frontier rather than optimizing either accuracy or speed alone. More details are in Appendix~\ref{additional experiment}.
\begin{wrapfigure}{r}{10.5cm}
\vspace{0pt}
\centering
\begin{minipage}[t]{0.49\linewidth}
\centering
\includegraphics[
    width=\linewidth,
    trim=8 8 8 8,
    clip
]{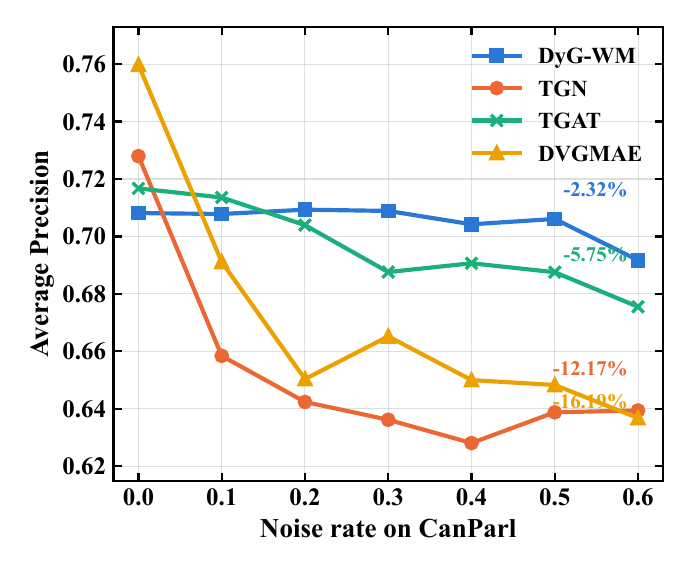}
\vspace{-4pt}
\end{minipage}
\hfill
\begin{minipage}[t]{0.49\linewidth}
\centering
\includegraphics[
    width=\linewidth,
    trim=8 8 8 8,
    clip
]{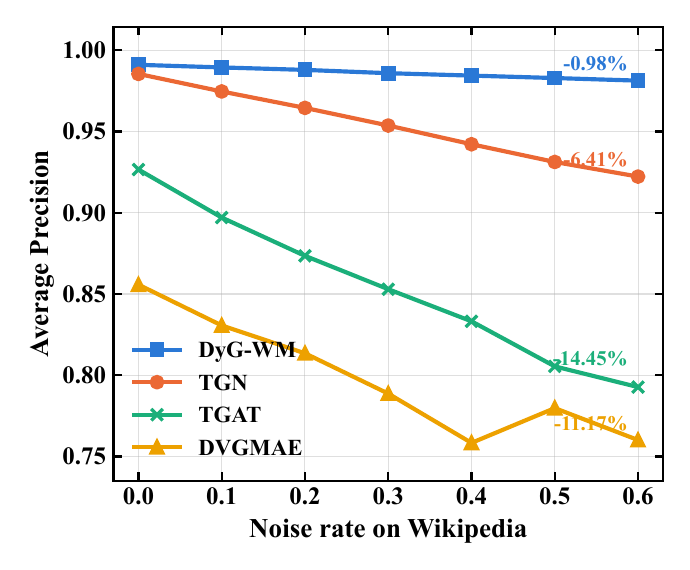}
\vspace{-4pt}
\end{minipage}
\vspace{-4pt}
\caption{Robustness to noise on Can.\ Parl. and Wikipedia. }
\vspace{-10pt}
\label{fig:robustness}
\end{wrapfigure}
\subsection{Robustness}
We train all models on clean data in the transductive setting with random negative sampling and add feature noise only at test time. For a noise rate $\sigma$, we randomly replace a fraction $\sigma$ of the input features with resampled values, with $\sigma$ ranging from $0.1$ to $0.6$. Figure~\ref{fig:robustness} shows that \method is less affected by increasing noise than the compared methods. On Wikipedia, its AP drops by only $0.98\%$ at $\sigma=0.6$, compared with $6.41\%$, $14.45\%$, and $11.17\%$ for TGN, TGAT, and DVGMAE. On Can.\ Parl., \method becomes the best-performing method from $\sigma=0.2$ onward and drops by only $2.32\%$ at $\sigma=0.6$.

\section{Conclusion}
We introduced \method, a dynamic graph world modelling framework for jointly modelling the evolution of nodes and relations. \method uses historical node-relation contexts and their ordered structural changes to predict future representations through a dual-level joint-embedding objective, without reconstructing the original graph observations. Experiments on 12 dynamic link prediction datasets and two dynamic node classification datasets show that the model performs well across both tasks. The ablation and latent-alignment results also show the contributions of the main components. Future work will consider more complex evolving relational systems and a broader range of downstream tasks.
\section*{AI Use Statement}
Generative AI tools were used solely for language polishing, grammar checking, and improving the clarity and readability of the manuscript. They were not used to develop the proposed methodology, design or conduct experiments, generate experimental results, or analyze and interpret the results. All AI-assisted revisions were carefully reviewed and verified by the authors. The authors take full responsibility for the final content of this work.

\bibliography{iclr2027_conference}
\bibliographystyle{iclr2027_conference}
\newpage
\appendix
\noindent{\large\textsc{Appendix}}

\noindent
This is the appendix of the work: ``\textbf{Dynamic Graph World Modelling: Learning Future Dynamics from Temporal Context}''. 
In this appendix, we provide additional methodological details, experimental settings, extended results, and further analyses of the proposed \method.
\section{Experimental Details}
\subsection{Datasets}
\label{app:datasets}
We evaluate link prediction on the twelve public dynamic graph datasets collected by
\cite{poursafaei2022towards} and redistributed with DyGLib~\citep{yu2023towards}; their
statistics are summarized in Table~\ref{tab:datasets}. Following SG-JEPA~\citep{zhang2026scalable}, we use the two large-scale datasets released by~\cite{lu2019temporal}, summarized in Table~\ref{tab:node_datasets}.

\begin{table}[h]
\centering
\footnotesize
\setlength{\tabcolsep}{3pt}
\caption{Statistics of the datasets. N/A denotes the absence of node or edge features.
\#Nodes denotes the number of nodes.}
\label{tab:datasets}
\resizebox{\textwidth}{!}{
\begin{tabular}{l|lrrcccrl}
\toprule
Datasets & Domains & \#Nodes & \#Links & \#N\&L Feature & Bipartite & Duration & Unique Steps & Time Granularity \\
\midrule
Wikipedia  & Social      & 9{,}227    & 157{,}474     & N/A \& 172 & True  & 1 month       & 152{,}757     & Unix timestamps \\
Reddit     & Social      & 10{,}984   & 672{,}447     & N/A \& 172 & True  & 1 month       & 669{,}065     & Unix timestamps \\
MOOC       & Interaction & 7{,}144    & 411{,}749     & N/A \& 4   & True  & 17 months     & 345{,}600     & Unix timestamps \\
LastFM     & Interaction & 1{,}980    & 1{,}293{,}103 & N/A \& N/A & True  & 1 month       & 1{,}283{,}614 & Unix timestamps \\
Enron      & Social      & 184        & 125{,}235     & N/A \& N/A & False & 3 years       & 22{,}632      & Unix timestamps \\
UCI        & Social      & 1{,}899    & 59{,}835      & N/A \& N/A & False & 196 days      & 58{,}911      & Unix timestamps \\
Can.\ Parl. & Politics    & 734        & 74{,}478      & N/A \& 1   & False & 14 years      & 14            & years \\
US Legis.  & Politics    & 225        & 60{,}396      & N/A \& 1   & False & 12 congresses & 12            & congresses \\
UN Trade   & Economics   & 255        & 507{,}497     & N/A \& 1   & False & 32 years      & 32            & years \\
UN Vote    & Politics    & 201        & 1{,}035{,}742 & N/A \& 1   & False & 72 years      & 72            & years \\
Contact    & Proximity   & 692        & 2{,}426{,}279 & N/A \& 1   & False & 1 month       & 8{,}064       & 5 minutes \\
Flights    & Transport   & 13{,}169   & 1{,}927{,}145 & N/A \& 1   & False & 4 months      & 122           & days \\
\bottomrule
\end{tabular}}
\end{table}

\begin{table}[h]
\centering
\small
\caption{Statistics of the dynamic node classification datasets.}
\label{tab:node_datasets}
\begin{tabular}{l|rrrcc}
\toprule
Datasets & \#Nodes & \#Links & \#Classes & \#Snapshots & \#Features \\
\midrule
DBLP   & 28{,}085      & 236{,}894      & 10 & 27   & 80 \\
Tmall  & 577{,}314     & 4{,}807{,}545  & 5  & 186 & 80 \\
\bottomrule
\end{tabular}
\end{table}

\subsection{Implementation Details}
\label{app:impl}

\noindent\textbf{Environment and Reproducibility.}
We implement all methods in Python using PyTorch. The code requires Python~3.10 or later and PyTorch~2.1 or later. All experiments are conducted on NVIDIA A100 GPUs.

\noindent\textbf{Shared Architecture Configuration.} Unless specified otherwise, both \method~and the snapshot-based baselines use a hidden dimension of 128 and two message-passing layers. For dynamic link prediction, we set $W=4$, corresponding to three historical snapshots in $\mathcal C_t$, with a 32-dimensional temporal encoding and a 256-dimensional predictor. For dynamic node classification, we set $W=3$, corresponding to two historical snapshots, with a 32-dimensional temporal encoding and a 256-dimensional predictor.

For \method, we use second-order path signatures throughout. In link prediction, the event-history representation and semantic projection have dimension 16, the relation-context budget is $B=32$, the temporal decay factor is $\lambda=0.8$, and the common-neighbor bonus is $\beta=1.0$. The target encoder uses $\eta_{\mathrm{ema}}=0.999$. The default loss weights are
$\lambda_N=0.02, \lambda_R=0.02, \lambda_{\mathrm{link}}=1.0, \lambda_{\mathrm{rank}}=1.0,
\lambda_{\mathrm{var}}=0.005, \lambda_{\mathrm{cov}}=0.0005$.
For node classification, the relation and link objectives are disabled, and we use
$\lambda_N=1.0, \lambda_{\mathrm{var}}=0.02, \lambda_{\mathrm{cov}}=0.001$,
with $\eta_{\mathrm{ema}}=0.995$ for the node target encoder.

\begin{table}[t]
    \centering
    \caption{Default implementation settings for \method. Dataset-specific overrides are selected using validation data only.}
    \label{tab:implementation-config}
    \small
    \begin{tabular}{lcc}
        \toprule
        Configuration & Link prediction & Node classification \\
        \midrule
        Hidden dimension & 128 & 128 \\
        Message-passing layers & 2 & 2 \\
        Window parameter $W$ & 4 & 3 \\
        Temporal dimension & 32 & 32 \\
        Predictor hidden dimension & 256 & 256 \\
        Signature order & 2 & 2 \\
        Relation-context budget $B$ & 32 & N/A \\
        EMA coefficient $\eta_{\mathrm{ema}}$ & 0.999 & 0.995 \\
        Maximum epochs & 100 & 100 \\
        Learning rate & $3\times10^{-4}$ & $10^{-3}$ \\
        Weight decay & $10^{-5}$ & $10^{-5}$ \\
        Gradient-norm threshold & 1.0 & 1.0 \\
        \bottomrule
    \end{tabular}
\end{table}

\noindent\textbf{Dynamic Link Prediction.}
We follow the unified DyGLib evaluation pipeline~\citep{yu2023towards}. Each dataset is split chronologically into 70\%, 15\%, and 15\% for training, validation, and testing. Under the default random-negative protocol, each observed interaction is paired with one uniformly corrupted destination during evaluation. Candidate collisions produced by the benchmark sampler are retained rather than deduplicated. Evaluation batches contain 200 positive interactions and their corresponding negatives. We use fixed negative-sampling seeds of 0 and 2 for validation and testing, respectively, and present the same sampled candidates to all methods.

We optimize \method~with AdamW using $\beta_1=0.9, \beta_2=0.95, \epsilon=10^{-10}$.
The default mini-batch contains 200 positive candidate groups and their one-to-one negatives, giving 400 candidate rows. We reduce the learning rate by a factor of $0.5$ after five validation evaluations without improvement, down to a minimum of $10^{-5}$. Models are evaluated after every epoch, and training stops after 20 evaluations without checkpoint improvement. Checkpoints are selected using validation AP and AUC only, and the test set is evaluated once using the selected checkpoint.

\noindent\textbf{Dynamic Node Classification.}
We train the node representation learner for at most 100 epochs using AdamW with
$ \beta_1=0.9, \beta_2=0.95, \epsilon=10^{-10}$,
and gradient clipping at 1.0. Node mini-batches contain 1,024 nodes for DBLP and Tmall. After representation learning, all encoders are frozen and an MLP probe with a hidden dimension of 512 is fitted to the final-snapshot representations. We use 40\%, 60\%, 80\% of the labelled nodes for probe training and reserve 10\% of this training portion for validation. Checkpoints are selected according to validation Macro-F1 after a minimum of 25 representation-learning epochs.

For \method, separate MLP probes are fitted to the predicted node representation and the online-encoder representation, and their logits are averaged for final classification. The target encoder is used only to construct representation targets during training and is not used by the downstream classifier.
\subsection{Detailed Evaluation Settings}
\label{app:evaluation-settings}

\noindent\textbf{Detailed Settings for Effectiveness Evaluation.}
To comprehensively evaluate the performance of all methods, we adopt three negative edge sampling strategies for dynamic link prediction following DyGLib~\citep{yu2023towards}. Let $\mathcal E_{\mathrm{train}}$ and $\mathcal E_{\mathrm{test}}$ denote the training and test edge sets, respectively. The edges can be divided into three categories: (a) edges appearing only during training, $\mathcal E_{\mathrm{train}}\setminus\mathcal E_{\mathrm{test}}$; (b) edges appearing in both training and testing, $\mathcal E_{\mathrm{train}}\cap\mathcal E_{\mathrm{test}}$; and (c) edges appearing only during testing, $\mathcal E_{\mathrm{test}}\setminus\mathcal E_{\mathrm{train}}$. We emphasize that inductive negative sampling is different from the new-node inductive setting. The three negative sampling strategies are described as follows:

\begin{itemize}
\item \textbf{Random Negative Sampling (\emph{rnd}).}
For each positive interaction, we retain its source node and timestamp while uniformly sampling a destination node from the complete destination set. The resulting source-destination pair is treated as a negative edge.

\item \textbf{Historical Negative Sampling (\emph{hist}).}
Historical negative sampling selects edges that appeared at earlier timestamps but are absent from the current evaluation batch. This setting evaluates whether a model can identify the specific time at which a previously observed edge reappears, rather than predicting that the edge remains active after its first occurrence. For an evaluation batch spanning timestamps $[t_s,t_e]$, negatives are sampled from edges observed strictly before $t_s$ after excluding edges appearing within $[t_s,t_e]$. If the historical candidate pool contains fewer edges than the number of positive examples, the remaining negatives are generated using random sampling.

\item \textbf{Inductive Negative Sampling (\emph{ind}).}
In contrast to historical sampling, inductive negative sampling focuses on edges that first appear during the evaluation period. After such an edge has appeared, the model is required to determine whether it will reoccur at subsequent timestamps. For each evaluation batch, we sample edges that were not observed before the current evaluation split, have appeared before the current batch, and are absent from the current batch. If the number of eligible inductive edges is insufficient, the remaining negatives are generated using random sampling.
\end{itemize}

\noindent\textbf{Detailed Settings for Robustness Evaluation.}
As shown in Figure~\ref{fig:robustness}, we evaluate model robustness to feature corruption on Wikipedia and Can.\ Parl. All models are trained on clean data under the transductive setting with random negative sampling, while noise is introduced only during evaluation. For a noise rate $\sigma$, we randomly select a fraction $\sigma$ of the input features and replace them with resampled values. We vary $\sigma$ from $0.1$ to $0.6$, with $\sigma=0$ denoting the clean setting, and use the same corruption seed for all compared methods at each noise level.

\subsection{Additional Experimental Results}
\label{additional experiment}
This subsection provides additional results that complement the main experiments. Tables~\ref{tab:ap_main} and~\ref{tab:ap_inductive} report Average Precision (AP) for dynamic link prediction under transductive and new-node inductive settings. We also provide additional latent-alignment visualizations on Wikipedia and UCI, together with the full accuracy-efficiency comparison across all 12 link prediction datasets in Figure~\ref{fig:efficiency_all}.
\begin{table*}[t]
\centering
\caption{\textbf{Transductive} Average Precision (AP) for dynamic link prediction with random (\emph{rnd}), historical (\emph{hist}), and inductive (\emph{ind}) negative edge sampling. \best{bold red} and \secondb{underlined blue} emphasize best and 2nd-best results.}
\label{tab:ap_main}
\resizebox{\textwidth}{!}{%
\begin{tabular}{llccccccccccc>{\columncolor{OursTint}}c}
\toprule
NSS & Datasets & JODIE & DyRep & TGAT & TGN & CAWN & TCL & GraphMixer & DyGFormer & CLDG & MaskDGNN & DVGMAE & \method \\
\midrule
\multirow{13}{*}{\emph{rnd}} & Wikipedia & 96.69{\scriptsize $\pm$ 0.21} & 95.20{\scriptsize $\pm$ 0.54} & 92.68{\scriptsize $\pm$ 0.47} & 98.54{\scriptsize $\pm$ 0.05} & 98.78{\scriptsize $\pm$ 0.01} & 97.93{\scriptsize $\pm$ 0.06} & 97.39{\scriptsize $\pm$ 0.05} & \secondb{99.06{\scriptsize $\pm$ 0.02}} & 91.55{\scriptsize $\pm$ 0.09} & 88.76{\scriptsize $\pm$ 0.43} & 85.68{\scriptsize $\pm$ 0.62} & \best{99.10{\scriptsize $\pm$ 0.02}} \\
 & Reddit & 98.07{\scriptsize $\pm$ 0.02} & 97.71{\scriptsize $\pm$ 0.56} & 98.03{\scriptsize $\pm$ 0.02} & 97.74{\scriptsize $\pm$ 0.39} & 99.18{\scriptsize $\pm$ 0.01} & 98.09{\scriptsize $\pm$ 0.04} & 98.09{\scriptsize $\pm$ 0.01} & \secondb{99.18{\scriptsize $\pm$ 0.00}} & 97.01{\scriptsize $\pm$ 0.02} & 96.97{\scriptsize $\pm$ 0.02} & 95.91{\scriptsize $\pm$ 0.07} & \best{99.28{\scriptsize $\pm$ 0.01}} \\
 & MOOC & 79.55{\scriptsize $\pm$ 2.76} & 75.48{\scriptsize $\pm$ 4.64} & 84.54{\scriptsize $\pm$ 0.05} & \best{92.81{\scriptsize $\pm$ 0.85}} & 81.50{\scriptsize $\pm$ 0.22} & 85.17{\scriptsize $\pm$ 0.49} & 83.79{\scriptsize $\pm$ 0.16} & \secondb{89.26{\scriptsize $\pm$ 0.26}} & 64.90{\scriptsize $\pm$ 1.88} & 72.35{\scriptsize $\pm$ 0.17} & 68.41{\scriptsize $\pm$ 1.18} & 81.94{\scriptsize $\pm$ 0.38} \\
 & LastFM & 71.67{\scriptsize $\pm$ 0.90} & 72.38{\scriptsize $\pm$ 0.75} & 73.39{\scriptsize $\pm$ 0.27} & 62.64{\scriptsize $\pm$ 2.26} & 86.99{\scriptsize $\pm$ 0.04} & 76.82{\scriptsize $\pm$ 0.38} & 75.81{\scriptsize $\pm$ 0.09} & \best{93.12{\scriptsize $\pm$ 0.05}} & 63.91{\scriptsize $\pm$ 0.63} & 65.76{\scriptsize $\pm$ 0.57} & 63.67{\scriptsize $\pm$ 0.19} & \secondb{92.20{\scriptsize $\pm$ 0.04}} \\
 & Enron & 84.06{\scriptsize $\pm$ 2.82} & 83.50{\scriptsize $\pm$ 2.64} & 76.08{\scriptsize $\pm$ 0.56} & 81.68{\scriptsize $\pm$ 2.21} & 89.42{\scriptsize $\pm$ 0.05} & 85.53{\scriptsize $\pm$ 0.30} & 81.62{\scriptsize $\pm$ 0.29} & \best{92.27{\scriptsize $\pm$ 0.09}} & 86.13{\scriptsize $\pm$ 0.35} & 84.57{\scriptsize $\pm$ 0.68} & 71.82{\scriptsize $\pm$ 0.58} & \secondb{91.40{\scriptsize $\pm$ 0.22}} \\
 & UCI & 89.16{\scriptsize $\pm$ 0.88} & 60.57{\scriptsize $\pm$ 8.46} & 82.60{\scriptsize $\pm$ 0.44} & 94.19{\scriptsize $\pm$ 0.38} & 95.17{\scriptsize $\pm$ 0.04} & 91.31{\scriptsize $\pm$ 3.18} & 93.51{\scriptsize $\pm$ 0.28} & \best{95.88{\scriptsize $\pm$ 0.14}} & 86.68{\scriptsize $\pm$ 0.14} & 87.71{\scriptsize $\pm$ 0.06} & 88.05{\scriptsize $\pm$ 0.10} & \secondb{95.26{\scriptsize $\pm$ 0.09}} \\
 & Flights & 90.89{\scriptsize $\pm$ 1.28} & 90.77{\scriptsize $\pm$ 0.97} & 94.44{\scriptsize $\pm$ 0.06} & 94.06{\scriptsize $\pm$ 0.59} & \secondb{98.51{\scriptsize $\pm$ 0.02}} & 91.40{\scriptsize $\pm$ 0.04} & 91.11{\scriptsize $\pm$ 0.02} & \best{98.94{\scriptsize $\pm$ 0.01}} & 95.00{\scriptsize $\pm$ 0.12} & 94.19{\scriptsize $\pm$ 0.06} & 92.04{\scriptsize $\pm$ 0.15} & 98.24{\scriptsize $\pm$ 0.10} \\
 & Can.\ Parl. & 72.25{\scriptsize $\pm$ 1.93} & 64.92{\scriptsize $\pm$ 0.94} & 70.45{\scriptsize $\pm$ 1.24} & 74.33{\scriptsize $\pm$ 1.48} & 69.91{\scriptsize $\pm$ 2.04} & 70.13{\scriptsize $\pm$ 0.96} & \secondb{77.46{\scriptsize $\pm$ 0.52}} & \best{97.18{\scriptsize $\pm$ 0.31}} & 70.14{\scriptsize $\pm$ 0.97} & 74.10{\scriptsize $\pm$ 1.50} & 76.06{\scriptsize $\pm$ 0.63} & 73.86{\scriptsize $\pm$ 0.68} \\
 & US Legis. & \best{77.97{\scriptsize $\pm$ 0.28}} & 70.93{\scriptsize $\pm$ 2.77} & 66.87{\scriptsize $\pm$ 0.53} & 75.92{\scriptsize $\pm$ 0.88} & 70.97{\scriptsize $\pm$ 0.16} & 69.24{\scriptsize $\pm$ 0.73} & 69.62{\scriptsize $\pm$ 0.69} & 70.40{\scriptsize $\pm$ 0.47} & 67.09{\scriptsize $\pm$ 1.68} & 74.97{\scriptsize $\pm$ 0.12} & 74.59{\scriptsize $\pm$ 0.28} & \secondb{76.81{\scriptsize $\pm$ 0.29}} \\
 & UN Trade & 64.76{\scriptsize $\pm$ 0.24} & 63.34{\scriptsize $\pm$ 0.32} & 64.16{\scriptsize $\pm$ 0.06} & 65.43{\scriptsize $\pm$ 0.62} & 65.77{\scriptsize $\pm$ 0.05} & 62.44{\scriptsize $\pm$ 0.02} & 63.51{\scriptsize $\pm$ 1.04} & 68.30{\scriptsize $\pm$ 2.28} & 65.85{\scriptsize $\pm$ 0.08} & 72.70{\scriptsize $\pm$ 0.29} & \secondb{73.46{\scriptsize $\pm$ 0.25}} & \best{76.75{\scriptsize $\pm$ 0.62}} \\
 & UN Vote & \secondb{65.45{\scriptsize $\pm$ 0.66}} & 64.84{\scriptsize $\pm$ 0.49} & 52.89{\scriptsize $\pm$ 0.27} & \best{67.60{\scriptsize $\pm$ 1.04}} & 53.06{\scriptsize $\pm$ 0.08} & 51.57{\scriptsize $\pm$ 0.23} & 52.08{\scriptsize $\pm$ 0.13} & 57.33{\scriptsize $\pm$ 0.24} & 50.86{\scriptsize $\pm$ 0.00} & 51.20{\scriptsize $\pm$ 0.81} & 53.03{\scriptsize $\pm$ 0.84} & 62.21{\scriptsize $\pm$ 0.17} \\
 & Contact & 95.27{\scriptsize $\pm$ 0.84} & 96.02{\scriptsize $\pm$ 0.17} & 96.62{\scriptsize $\pm$ 0.04} & 95.60{\scriptsize $\pm$ 0.97} & 90.66{\scriptsize $\pm$ 0.13} & 94.67{\scriptsize $\pm$ 0.19} & 91.81{\scriptsize $\pm$ 0.08} & \best{98.34{\scriptsize $\pm$ 0.01}} & 70.93{\scriptsize $\pm$ 0.38} & 74.53{\scriptsize $\pm$ 0.43} & 69.61{\scriptsize $\pm$ 0.32} & \secondb{98.06{\scriptsize $\pm$ 0.04}} \\
\cmidrule(l){2-14}
\rowcolor{RankGray}
 & Avg.\ Rank & 6.50 & 8.50 & 7.75 & 5.42 & 5.17 & 7.17 & 6.92 & \best{2.42} & 9.00 & 7.92 & 8.58 & \secondb{2.67} \\
\midrule
\multirow{13}{*}{\emph{hist}} & Wikipedia & 84.19{\scriptsize $\pm$ 0.52} & 79.75{\scriptsize $\pm$ 0.61} & 71.19{\scriptsize $\pm$ 0.87} & \secondb{86.98{\scriptsize $\pm$ 0.27}} & 68.24{\scriptsize $\pm$ 7.35} & 86.77{\scriptsize $\pm$ 3.35} & \best{90.99{\scriptsize $\pm$ 0.33}} & 84.55{\scriptsize $\pm$ 2.14} & 77.21{\scriptsize $\pm$ 0.14} & 75.87{\scriptsize $\pm$ 0.66} & 74.48{\scriptsize $\pm$ 0.69} & 84.06{\scriptsize $\pm$ 0.93} \\
 & Reddit & 79.23{\scriptsize $\pm$ 0.63} & 77.96{\scriptsize $\pm$ 2.11} & 77.20{\scriptsize $\pm$ 0.12} & 77.92{\scriptsize $\pm$ 1.17} & 80.85{\scriptsize $\pm$ 0.25} & 80.81{\scriptsize $\pm$ 0.54} & 78.77{\scriptsize $\pm$ 0.24} & \secondb{81.12{\scriptsize $\pm$ 0.90}} & 76.12{\scriptsize $\pm$ 0.26} & 78.35{\scriptsize $\pm$ 0.21} & 75.93{\scriptsize $\pm$ 0.38} & \best{83.84{\scriptsize $\pm$ 0.62}} \\
 & MOOC & 76.75{\scriptsize $\pm$ 4.57} & 67.32{\scriptsize $\pm$ 4.48} & 82.40{\scriptsize $\pm$ 0.61} & \best{89.38{\scriptsize $\pm$ 2.13}} & 74.63{\scriptsize $\pm$ 0.85} & 80.86{\scriptsize $\pm$ 0.99} & 83.33{\scriptsize $\pm$ 1.67} & \secondb{88.27{\scriptsize $\pm$ 1.02}} & 59.34{\scriptsize $\pm$ 2.39} & 71.43{\scriptsize $\pm$ 1.40} & 69.74{\scriptsize $\pm$ 2.05} & 81.52{\scriptsize $\pm$ 2.35} \\
 & LastFM & 76.15{\scriptsize $\pm$ 2.39} & 71.65{\scriptsize $\pm$ 2.46} & 70.70{\scriptsize $\pm$ 0.34} & 59.74{\scriptsize $\pm$ 6.41} & 69.68{\scriptsize $\pm$ 0.11} & 71.57{\scriptsize $\pm$ 1.41} & 72.08{\scriptsize $\pm$ 0.39} & \secondb{81.55{\scriptsize $\pm$ 0.33}} & 59.78{\scriptsize $\pm$ 0.86} & 61.99{\scriptsize $\pm$ 0.76} & 61.39{\scriptsize $\pm$ 1.17} & \best{81.95{\scriptsize $\pm$ 0.61}} \\
 & Enron & 67.04{\scriptsize $\pm$ 2.26} & 73.25{\scriptsize $\pm$ 1.48} & 64.87{\scriptsize $\pm$ 0.37} & 70.50{\scriptsize $\pm$ 2.50} & 65.56{\scriptsize $\pm$ 0.18} & 71.66{\scriptsize $\pm$ 0.81} & \best{78.25{\scriptsize $\pm$ 0.29}} & 75.92{\scriptsize $\pm$ 0.49} & 68.89{\scriptsize $\pm$ 1.24} & 65.65{\scriptsize $\pm$ 1.40} & 61.39{\scriptsize $\pm$ 2.23} & \secondb{76.82{\scriptsize $\pm$ 1.87}} \\
 & UCI & 79.81{\scriptsize $\pm$ 2.74} & 52.95{\scriptsize $\pm$ 6.43} & 69.36{\scriptsize $\pm$ 0.43} & 81.47{\scriptsize $\pm$ 1.33} & 65.50{\scriptsize $\pm$ 0.45} & \secondb{81.54{\scriptsize $\pm$ 5.46}} & \best{84.70{\scriptsize $\pm$ 0.45}} & 81.39{\scriptsize $\pm$ 0.43} & 72.67{\scriptsize $\pm$ 0.34} & 74.60{\scriptsize $\pm$ 0.91} & 76.21{\scriptsize $\pm$ 0.54} & 74.38{\scriptsize $\pm$ 0.84} \\
 & Flights & 63.22{\scriptsize $\pm$ 0.77} & 59.23{\scriptsize $\pm$ 2.13} & 71.87{\scriptsize $\pm$ 0.23} & 67.80{\scriptsize $\pm$ 2.30} & 64.82{\scriptsize $\pm$ 0.96} & 70.89{\scriptsize $\pm$ 0.23} & 71.76{\scriptsize $\pm$ 0.22} & 64.93{\scriptsize $\pm$ 1.39} & \secondb{72.90{\scriptsize $\pm$ 0.35}} & \best{73.48{\scriptsize $\pm$ 0.40}} & 70.91{\scriptsize $\pm$ 0.78} & 60.98{\scriptsize $\pm$ 2.16} \\
 & Can.\ Parl. & 67.77{\scriptsize $\pm$ 12.02} & 62.85{\scriptsize $\pm$ 1.28} & 66.12{\scriptsize $\pm$ 2.02} & 72.39{\scriptsize $\pm$ 3.46} & 66.24{\scriptsize $\pm$ 2.27} & 67.23{\scriptsize $\pm$ 1.65} & 77.43{\scriptsize $\pm$ 1.52} & \best{97.46{\scriptsize $\pm$ 0.63}} & 77.24{\scriptsize $\pm$ 1.44} & 82.21{\scriptsize $\pm$ 2.40} & \secondb{82.26{\scriptsize $\pm$ 0.71}} & 73.28{\scriptsize $\pm$ 1.94} \\
 & US Legis. & 64.99{\scriptsize $\pm$ 4.62} & 76.62{\scriptsize $\pm$ 3.77} & 52.92{\scriptsize $\pm$ 1.97} & 66.33{\scriptsize $\pm$ 5.57} & 68.96{\scriptsize $\pm$ 8.28} & 69.98{\scriptsize $\pm$ 7.32} & 62.89{\scriptsize $\pm$ 10.19} & 66.57{\scriptsize $\pm$ 8.52} & 73.81{\scriptsize $\pm$ 4.42} & 83.93{\scriptsize $\pm$ 6.04} & \best{87.88{\scriptsize $\pm$ 2.05}} & \secondb{84.55{\scriptsize $\pm$ 2.56}} \\
 & UN Trade & 55.13{\scriptsize $\pm$ 0.64} & 57.06{\scriptsize $\pm$ 0.86} & 57.55{\scriptsize $\pm$ 0.42} & 58.89{\scriptsize $\pm$ 1.04} & 58.32{\scriptsize $\pm$ 0.36} & 55.64{\scriptsize $\pm$ 0.40} & 59.34{\scriptsize $\pm$ 3.41} & 65.91{\scriptsize $\pm$ 4.09} & 61.87{\scriptsize $\pm$ 1.19} & \secondb{72.20{\scriptsize $\pm$ 1.41}} & 72.02{\scriptsize $\pm$ 1.91} & \best{73.20{\scriptsize $\pm$ 2.07}} \\
 & UN Vote & \secondb{72.85{\scriptsize $\pm$ 0.94}} & 70.41{\scriptsize $\pm$ 0.89} & 55.99{\scriptsize $\pm$ 0.87} & 70.57{\scriptsize $\pm$ 1.12} & 51.45{\scriptsize $\pm$ 0.34} & 51.43{\scriptsize $\pm$ 1.04} & 52.07{\scriptsize $\pm$ 1.16} & 61.66{\scriptsize $\pm$ 0.66} & 50.57{\scriptsize $\pm$ 0.00} & 53.57{\scriptsize $\pm$ 3.00} & 56.37{\scriptsize $\pm$ 0.93} & \best{76.68{\scriptsize $\pm$ 4.03}} \\
 & Contact & 95.21{\scriptsize $\pm$ 1.09} & 96.22{\scriptsize $\pm$ 0.29} & 96.53{\scriptsize $\pm$ 0.24} & 93.81{\scriptsize $\pm$ 2.71} & 84.67{\scriptsize $\pm$ 0.39} & 95.33{\scriptsize $\pm$ 0.23} & 93.55{\scriptsize $\pm$ 0.28} & \best{97.52{\scriptsize $\pm$ 0.18}} & 63.62{\scriptsize $\pm$ 0.77} & 69.80{\scriptsize $\pm$ 0.30} & 64.20{\scriptsize $\pm$ 0.82} & \secondb{97.16{\scriptsize $\pm$ 0.11}} \\
\cmidrule(l){2-14}
\rowcolor{RankGray}
 & Avg.\ Rank & 6.75 & 7.75 & 8.17 & 6.08 & 8.75 & 6.17 & 4.83 & \best{3.67} & 8.25 & 6.42 & 7.33 & \secondb{3.83} \\
\midrule
\multirow{13}{*}{\emph{ind}} & Wikipedia & 77.53{\scriptsize $\pm$ 0.77} & 73.50{\scriptsize $\pm$ 1.05} & 74.27{\scriptsize $\pm$ 0.84} & \secondb{85.92{\scriptsize $\pm$ 0.56}} & 70.11{\scriptsize $\pm$ 8.39} & 74.74{\scriptsize $\pm$ 6.48} & \best{88.87{\scriptsize $\pm$ 0.32}} & 83.09{\scriptsize $\pm$ 4.38} & 64.95{\scriptsize $\pm$ 0.54} & 63.25{\scriptsize $\pm$ 0.79} & 62.64{\scriptsize $\pm$ 0.94} & 82.28{\scriptsize $\pm$ 3.02} \\
 & Reddit & 86.44{\scriptsize $\pm$ 0.25} & 85.14{\scriptsize $\pm$ 2.15} & 87.77{\scriptsize $\pm$ 0.11} & 84.88{\scriptsize $\pm$ 1.65} & \best{91.98{\scriptsize $\pm$ 0.13}} & 79.43{\scriptsize $\pm$ 1.21} & 87.55{\scriptsize $\pm$ 0.07} & 91.14{\scriptsize $\pm$ 0.38} & 75.45{\scriptsize $\pm$ 0.97} & 80.31{\scriptsize $\pm$ 0.62} & 80.97{\scriptsize $\pm$ 0.23} & \secondb{91.92{\scriptsize $\pm$ 0.30}} \\
 & MOOC & 67.62{\scriptsize $\pm$ 2.01} & 58.75{\scriptsize $\pm$ 3.92} & 77.44{\scriptsize $\pm$ 1.01} & \secondb{82.50{\scriptsize $\pm$ 3.35}} & 74.29{\scriptsize $\pm$ 0.65} & 77.91{\scriptsize $\pm$ 0.90} & 77.34{\scriptsize $\pm$ 1.25} & \best{83.00{\scriptsize $\pm$ 0.66}} & 53.96{\scriptsize $\pm$ 2.63} & 56.99{\scriptsize $\pm$ 0.87} & 57.56{\scriptsize $\pm$ 0.29} & 77.10{\scriptsize $\pm$ 3.38} \\
 & LastFM & 65.37{\scriptsize $\pm$ 2.09} & 63.86{\scriptsize $\pm$ 1.77} & 70.46{\scriptsize $\pm$ 0.31} & 57.04{\scriptsize $\pm$ 2.80} & 67.21{\scriptsize $\pm$ 0.18} & 63.12{\scriptsize $\pm$ 1.54} & 68.11{\scriptsize $\pm$ 0.14} & \secondb{73.21{\scriptsize $\pm$ 1.20}} & 52.23{\scriptsize $\pm$ 0.48} & 52.83{\scriptsize $\pm$ 0.51} & 51.97{\scriptsize $\pm$ 0.12} & \best{75.73{\scriptsize $\pm$ 2.12}} \\
 & Enron & 70.43{\scriptsize $\pm$ 2.10} & 70.56{\scriptsize $\pm$ 2.41} & 65.76{\scriptsize $\pm$ 0.32} & 70.24{\scriptsize $\pm$ 3.19} & 76.63{\scriptsize $\pm$ 0.27} & 71.52{\scriptsize $\pm$ 0.39} & 74.79{\scriptsize $\pm$ 0.27} & \secondb{77.77{\scriptsize $\pm$ 0.99}} & 66.57{\scriptsize $\pm$ 1.18} & 66.84{\scriptsize $\pm$ 1.38} & 60.25{\scriptsize $\pm$ 2.38} & \best{83.28{\scriptsize $\pm$ 0.94}} \\
 & UCI & 64.41{\scriptsize $\pm$ 1.65} & 52.57{\scriptsize $\pm$ 1.43} & 68.44{\scriptsize $\pm$ 0.34} & 70.27{\scriptsize $\pm$ 0.77} & 64.61{\scriptsize $\pm$ 0.63} & \secondb{76.11{\scriptsize $\pm$ 2.59}} & \best{80.05{\scriptsize $\pm$ 0.39}} & 72.41{\scriptsize $\pm$ 0.55} & 52.39{\scriptsize $\pm$ 0.50} & 55.07{\scriptsize $\pm$ 0.31} & 55.84{\scriptsize $\pm$ 0.16} & 65.38{\scriptsize $\pm$ 1.11} \\
 & Flights & 65.43{\scriptsize $\pm$ 1.59} & 60.92{\scriptsize $\pm$ 2.63} & 74.35{\scriptsize $\pm$ 0.50} & 72.31{\scriptsize $\pm$ 2.95} & 69.33{\scriptsize $\pm$ 0.91} & 75.07{\scriptsize $\pm$ 0.23} & 75.14{\scriptsize $\pm$ 0.26} & 68.16{\scriptsize $\pm$ 1.79} & \secondb{75.85{\scriptsize $\pm$ 0.44}} & \best{78.06{\scriptsize $\pm$ 0.96}} & 74.99{\scriptsize $\pm$ 0.72} & 66.84{\scriptsize $\pm$ 0.95} \\
 & Can.\ Parl. & 62.31{\scriptsize $\pm$ 11.75} & 59.72{\scriptsize $\pm$ 0.67} & 65.90{\scriptsize $\pm$ 1.24} & 70.18{\scriptsize $\pm$ 4.04} & 68.44{\scriptsize $\pm$ 1.10} & 68.13{\scriptsize $\pm$ 1.14} & 71.20{\scriptsize $\pm$ 1.10} & \best{96.96{\scriptsize $\pm$ 0.39}} & 74.72{\scriptsize $\pm$ 1.21} & 79.03{\scriptsize $\pm$ 0.86} & \secondb{80.30{\scriptsize $\pm$ 0.30}} & 71.60{\scriptsize $\pm$ 2.66} \\
 & US Legis. & 67.60{\scriptsize $\pm$ 4.48} & 75.69{\scriptsize $\pm$ 3.52} & 51.80{\scriptsize $\pm$ 2.26} & 61.28{\scriptsize $\pm$ 5.31} & 64.51{\scriptsize $\pm$ 6.78} & 69.09{\scriptsize $\pm$ 10.16} & 61.38{\scriptsize $\pm$ 10.93} & 63.99{\scriptsize $\pm$ 7.41} & 73.48{\scriptsize $\pm$ 2.09} & 83.60{\scriptsize $\pm$ 2.11} & \best{87.49{\scriptsize $\pm$ 0.11}} & \secondb{85.47{\scriptsize $\pm$ 0.93}} \\
 & UN Trade & 58.21{\scriptsize $\pm$ 0.90} & 58.41{\scriptsize $\pm$ 0.62} & 63.50{\scriptsize $\pm$ 0.45} & 60.83{\scriptsize $\pm$ 1.84} & 65.28{\scriptsize $\pm$ 0.48} & 60.72{\scriptsize $\pm$ 0.49} & 62.56{\scriptsize $\pm$ 3.46} & 66.78{\scriptsize $\pm$ 5.38} & 62.42{\scriptsize $\pm$ 0.81} & 72.69{\scriptsize $\pm$ 1.17} & \secondb{72.91{\scriptsize $\pm$ 0.72}} & \best{78.71{\scriptsize $\pm$ 0.99}} \\
 & UN Vote & \secondb{70.10{\scriptsize $\pm$ 1.17}} & 69.01{\scriptsize $\pm$ 1.08} & 54.47{\scriptsize $\pm$ 0.68} & 69.11{\scriptsize $\pm$ 1.16} & 53.97{\scriptsize $\pm$ 1.11} & 51.70{\scriptsize $\pm$ 1.37} & 50.98{\scriptsize $\pm$ 0.39} & 53.46{\scriptsize $\pm$ 0.22} & 50.57{\scriptsize $\pm$ 0.00} & 51.39{\scriptsize $\pm$ 1.62} & 55.86{\scriptsize $\pm$ 1.43} & \best{78.42{\scriptsize $\pm$ 0.85}} \\
 & Contact & 93.46{\scriptsize $\pm$ 1.18} & 94.69{\scriptsize $\pm$ 0.45} & \secondb{95.37{\scriptsize $\pm$ 0.28}} & 91.31{\scriptsize $\pm$ 3.33} & 90.62{\scriptsize $\pm$ 0.37} & 92.76{\scriptsize $\pm$ 0.33} & 91.33{\scriptsize $\pm$ 0.28} & 95.13{\scriptsize $\pm$ 0.60} & 56.90{\scriptsize $\pm$ 0.83} & 61.76{\scriptsize $\pm$ 1.33} & 55.20{\scriptsize $\pm$ 0.73} & \best{96.99{\scriptsize $\pm$ 0.31}} \\
\cmidrule(l){2-14}
\rowcolor{RankGray}
 & Avg.\ Rank & 7.33 & 7.92 & 6.33 & 6.50 & 6.42 & 6.58 & 5.33 & \secondb{4.00} & 9.08 & 7.58 & 7.58 & \best{3.33} \\
\bottomrule
\end{tabular}%
}
\end{table*}

\begin{table*}[h]
\centering
\caption{\textbf{Inductive} Average Precision (AP) for dynamic link prediction in the inductive (new-node) setting with random negative edge sampling. \best{Bold red} and \secondb{underlined blue} indicate the best and second-best results, respectively.}
\label{tab:ap_inductive}
\resizebox{\textwidth}{!}{%
\begin{tabular}{lccccccccccc>{\columncolor{OursTint}}c}
\toprule
Datasets & JODIE & DyRep & TGAT & TGN & CAWN & TCL & GraphMixer & DyGFormer & CLDG & MaskDGNN & DVGMAE & \method \\
\midrule

Wikipedia
& 94.75{\scriptsize $\pm$ 0.20}
& 92.25{\scriptsize $\pm$ 0.22}
& 80.33{\scriptsize $\pm$ 0.69}
& 97.45{\scriptsize $\pm$ 0.06}
& 97.22{\scriptsize $\pm$ 0.03}
& 97.26{\scriptsize $\pm$ 0.08}
& 96.33{\scriptsize $\pm$ 0.05}
& \secondb{97.56{\scriptsize $\pm$ 0.03}}
& 89.39{\scriptsize $\pm$ 0.28}
& 86.59{\scriptsize $\pm$ 0.20}
& 82.19{\scriptsize $\pm$ 1.04}
& \best{97.94{\scriptsize $\pm$ 0.02}} \\

Reddit
& 95.87{\scriptsize $\pm$ 0.13}
& 95.19{\scriptsize $\pm$ 0.23}
& 66.20{\scriptsize $\pm$ 1.64}
& 96.24{\scriptsize $\pm$ 0.66}
& 97.66{\scriptsize $\pm$ 0.03}
& 96.45{\scriptsize $\pm$ 0.06}
& 95.61{\scriptsize $\pm$ 0.08}
& \secondb{97.74{\scriptsize $\pm$ 0.02}}
& 94.68{\scriptsize $\pm$ 0.06}
& 95.02{\scriptsize $\pm$ 0.04}
& 93.54{\scriptsize $\pm$ 0.18}
& \best{97.90{\scriptsize $\pm$ 0.06}} \\

MOOC
& 78.67{\scriptsize $\pm$ 1.16}
& 75.82{\scriptsize $\pm$ 3.14}
& 79.46{\scriptsize $\pm$ 0.43}
& \best{89.71{\scriptsize $\pm$ 0.64}}
& 81.86{\scriptsize $\pm$ 0.21}
& 81.88{\scriptsize $\pm$ 1.08}
& 81.71{\scriptsize $\pm$ 0.20}
& \secondb{87.24{\scriptsize $\pm$ 0.30}}
& 61.43{\scriptsize $\pm$ 0.51}
& 69.48{\scriptsize $\pm$ 0.39}
& 65.55{\scriptsize $\pm$ 0.93}
& 79.74{\scriptsize $\pm$ 0.48} \\

LastFM
& 82.84{\scriptsize $\pm$ 0.84}
& 82.01{\scriptsize $\pm$ 1.45}
& 78.54{\scriptsize $\pm$ 0.16}
& 62.93{\scriptsize $\pm$ 3.73}
& 89.44{\scriptsize $\pm$ 0.04}
& 80.93{\scriptsize $\pm$ 0.78}
& 81.98{\scriptsize $\pm$ 0.11}
& \secondb{92.22{\scriptsize $\pm$ 0.09}}
& 66.60{\scriptsize $\pm$ 1.18}
& 67.66{\scriptsize $\pm$ 0.63}
& 65.99{\scriptsize $\pm$ 0.41}
& \best{92.61{\scriptsize $\pm$ 0.17}} \\

Enron
& 80.62{\scriptsize $\pm$ 2.01}
& 70.73{\scriptsize $\pm$ 1.73}
& 71.10{\scriptsize $\pm$ 0.35}
& 72.08{\scriptsize $\pm$ 2.32}
& 86.31{\scriptsize $\pm$ 0.21}
& 82.54{\scriptsize $\pm$ 0.40}
& 76.70{\scriptsize $\pm$ 0.16}
& \best{89.89{\scriptsize $\pm$ 0.13}}
& 76.55{\scriptsize $\pm$ 1.20}
& 74.24{\scriptsize $\pm$ 1.44}
& 59.32{\scriptsize $\pm$ 0.79}
& \secondb{86.75{\scriptsize $\pm$ 0.56}} \\

UCI
& 78.46{\scriptsize $\pm$ 3.21}
& 58.76{\scriptsize $\pm$ 6.47}
& 79.51{\scriptsize $\pm$ 0.30}
& 87.78{\scriptsize $\pm$ 2.28}
& 92.10{\scriptsize $\pm$ 0.09}
& 82.22{\scriptsize $\pm$ 1.31}
& 90.91{\scriptsize $\pm$ 0.13}
& \secondb{92.99{\scriptsize $\pm$ 0.06}}
& 76.78{\scriptsize $\pm$ 0.32}
& 77.92{\scriptsize $\pm$ 0.24}
& 77.93{\scriptsize $\pm$ 0.15}
& \best{93.38{\scriptsize $\pm$ 0.18}} \\

Flights
& 90.27{\scriptsize $\pm$ 1.51}
& 86.50{\scriptsize $\pm$ 1.20}
& 87.76{\scriptsize $\pm$ 0.12}
& 86.35{\scriptsize $\pm$ 0.98}
& \secondb{97.32{\scriptsize $\pm$ 0.02}}
& 84.89{\scriptsize $\pm$ 0.04}
& 84.37{\scriptsize $\pm$ 0.19}
& \best{98.02{\scriptsize $\pm$ 0.01}}
& 90.94{\scriptsize $\pm$ 0.24}
& 89.79{\scriptsize $\pm$ 0.09}
& 85.67{\scriptsize $\pm$ 0.39}
& 94.14{\scriptsize $\pm$ 0.64} \\

Can.\ Parl.
& 54.06{\scriptsize $\pm$ 2.10}
& 55.36{\scriptsize $\pm$ 1.35}
& 57.28{\scriptsize $\pm$ 0.95}
& 58.74{\scriptsize $\pm$ 1.04}
& 59.72{\scriptsize $\pm$ 0.20}
& 58.41{\scriptsize $\pm$ 0.30}
& 57.82{\scriptsize $\pm$ 0.58}
& \best{95.58{\scriptsize $\pm$ 0.45}}
& 61.62{\scriptsize $\pm$ 0.70}
& 56.60{\scriptsize $\pm$ 1.22}
& 60.20{\scriptsize $\pm$ 0.83}
& \secondb{61.77{\scriptsize $\pm$ 0.63}} \\

US Legis.
& \secondb{61.14{\scriptsize $\pm$ 0.31}}
& 57.16{\scriptsize $\pm$ 0.76}
& 49.86{\scriptsize $\pm$ 0.87}
& 59.52{\scriptsize $\pm$ 0.49}
& 55.37{\scriptsize $\pm$ 0.98}
& 52.15{\scriptsize $\pm$ 0.77}
& 53.46{\scriptsize $\pm$ 1.42}
& 55.03{\scriptsize $\pm$ 0.11}
& 56.90{\scriptsize $\pm$ 3.39}
& 60.28{\scriptsize $\pm$ 0.47}
& 60.47{\scriptsize $\pm$ 0.41}
& \best{61.66{\scriptsize $\pm$ 1.99}} \\

UN Trade
& 58.60{\scriptsize $\pm$ 0.49}
& 56.85{\scriptsize $\pm$ 1.42}
& 55.40{\scriptsize $\pm$ 1.77}
& 60.02{\scriptsize $\pm$ 0.86}
& 65.12{\scriptsize $\pm$ 0.20}
& 62.09{\scriptsize $\pm$ 0.10}
& 63.49{\scriptsize $\pm$ 0.89}
& 70.05{\scriptsize $\pm$ 3.61}
& 65.72{\scriptsize $\pm$ 0.25}
& \secondb{72.18{\scriptsize $\pm$ 0.13}}
& \best{73.33{\scriptsize $\pm$ 0.27}}
& 62.74{\scriptsize $\pm$ 1.48} \\

UN Vote
& \secondb{57.33{\scriptsize $\pm$ 0.92}}
& 57.13{\scriptsize $\pm$ 0.83}
& 51.28{\scriptsize $\pm$ 0.46}
& \best{59.57{\scriptsize $\pm$ 0.37}}
& 50.30{\scriptsize $\pm$ 0.32}
& 50.64{\scriptsize $\pm$ 0.88}
& 51.08{\scriptsize $\pm$ 0.34}
& 55.22{\scriptsize $\pm$ 0.34}
& 50.98{\scriptsize $\pm$ 0.12}
& 49.18{\scriptsize $\pm$ 0.53}
& 50.62{\scriptsize $\pm$ 0.48}
& 55.14{\scriptsize $\pm$ 1.17} \\

Contact
& 95.11{\scriptsize $\pm$ 1.04}
& 92.58{\scriptsize $\pm$ 0.38}
& 95.61{\scriptsize $\pm$ 0.09}
& 92.97{\scriptsize $\pm$ 1.26}
& 89.88{\scriptsize $\pm$ 0.18}
& 93.87{\scriptsize $\pm$ 0.19}
& 90.36{\scriptsize $\pm$ 0.05}
& \secondb{96.03{\scriptsize $\pm$ 0.01}}
& 68.19{\scriptsize $\pm$ 0.95}
& 70.30{\scriptsize $\pm$ 0.66}
& 65.54{\scriptsize $\pm$ 0.46}
& \best{96.89{\scriptsize $\pm$ 0.05}} \\

\cmidrule(l){1-13}
\rowcolor{RankGray}
Avg.\ Rank
& 6.08
& 8.25
& 8.75
& 5.92
& \secondb{5.08}
& 6.58
& 7.08
& \best{2.58}
& 8.00
& 8.33
& 8.75
& \best{2.58} \\

\bottomrule
\end{tabular}%
}
\end{table*}
\begin{figure}
    \centering
    \includegraphics[width=\textwidth]{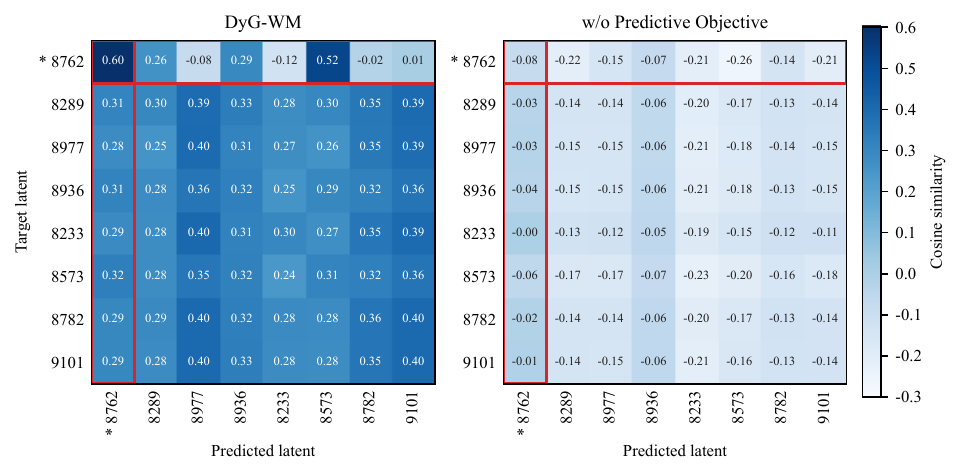}
    \vspace{-20pt}
    \caption{Latent alignment with and without the predictive objective in the Wikipedia dataset.}
    \label{fig:ablation_heatmap}
    \vspace{-20pt}
\end{figure}

\begin{figure}
    \centering
    \includegraphics[width=\textwidth]{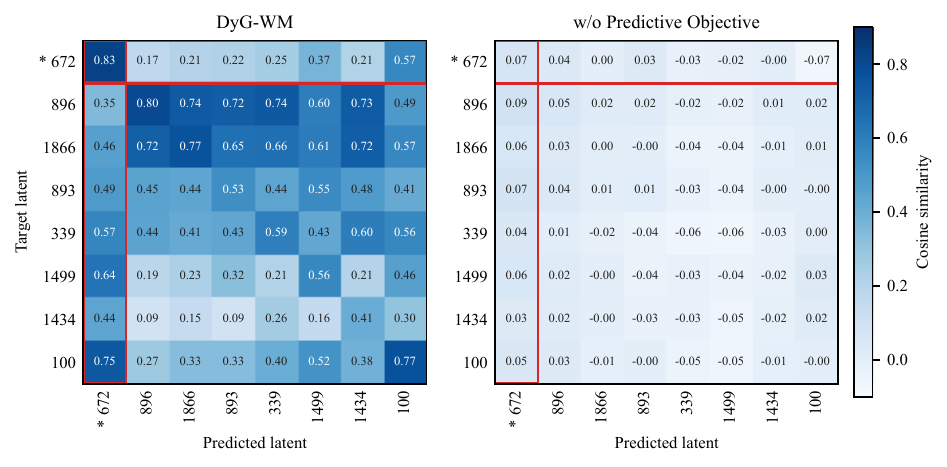}
    \vspace{-20pt}
    \caption{Latent alignment with and without the predictive objective in the UCI dataset.}
    \label{fig:ablation_heatmap}
    \vspace{-20pt}
\end{figure}

\begin{figure*}[h]
\centering
\begin{minipage}[t]{0.24\textwidth}
\centering
\includegraphics[width=\linewidth]{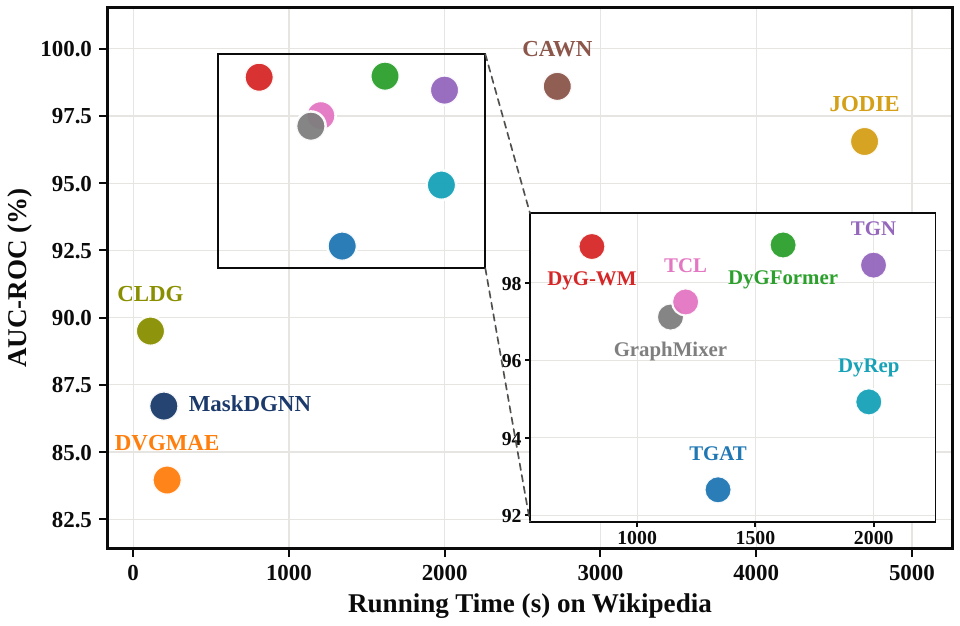}
{\scriptsize (a) Wikipedia}
\end{minipage}
\hfill
\begin{minipage}[t]{0.24\textwidth}
\centering
\includegraphics[width=\linewidth]{figure/efficiency_reddit.pdf}
\vspace{-3pt}
{\scriptsize (b) Reddit}
\end{minipage}
\hfill
\begin{minipage}[t]{0.24\textwidth}
\centering
\includegraphics[width=\linewidth]{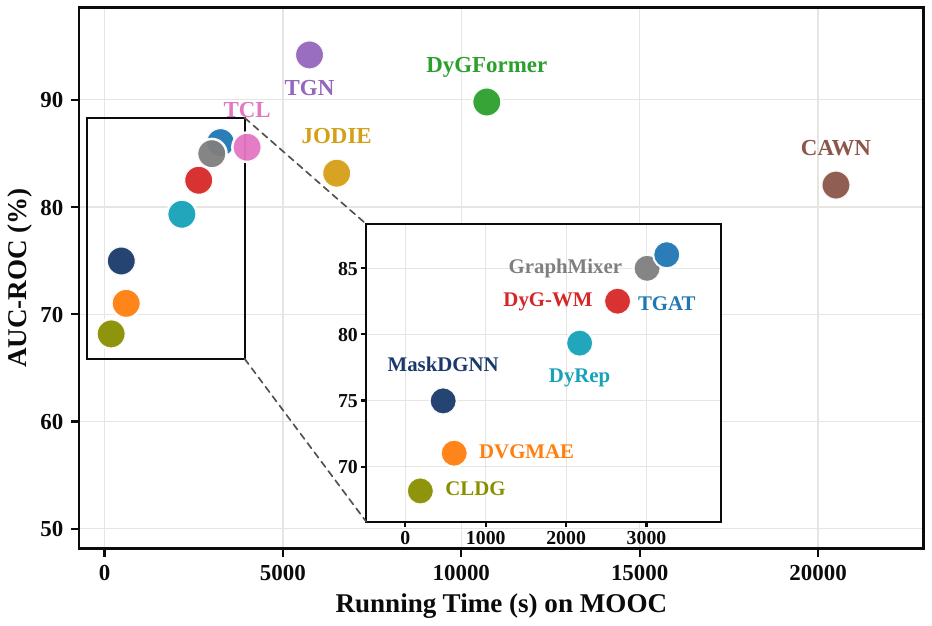}
\vspace{-3pt}
{\scriptsize (c) MOOC}
\end{minipage}
\hfill
\begin{minipage}[t]{0.24\textwidth}
\centering
\includegraphics[width=\linewidth]{figure/efficiency_lastfm.pdf}
\vspace{-3pt}
{\scriptsize (d) LastFM}
\end{minipage}
\begin{minipage}[t]{0.24\textwidth}
\centering
\includegraphics[width=\linewidth]{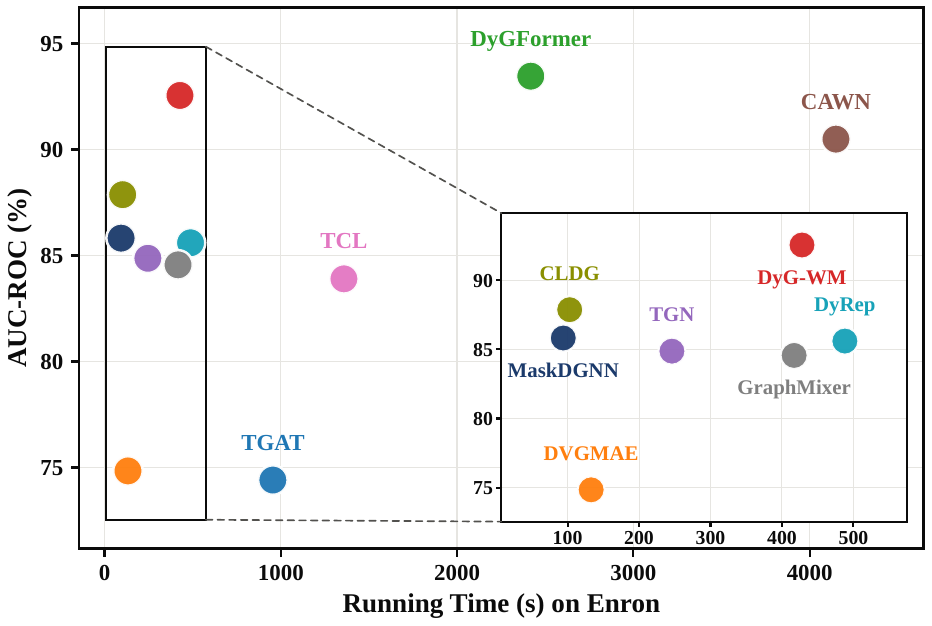}
{\scriptsize (e) Enron}
\end{minipage}
\hfill
\begin{minipage}[t]{0.24\textwidth}
\centering
\includegraphics[width=\linewidth]{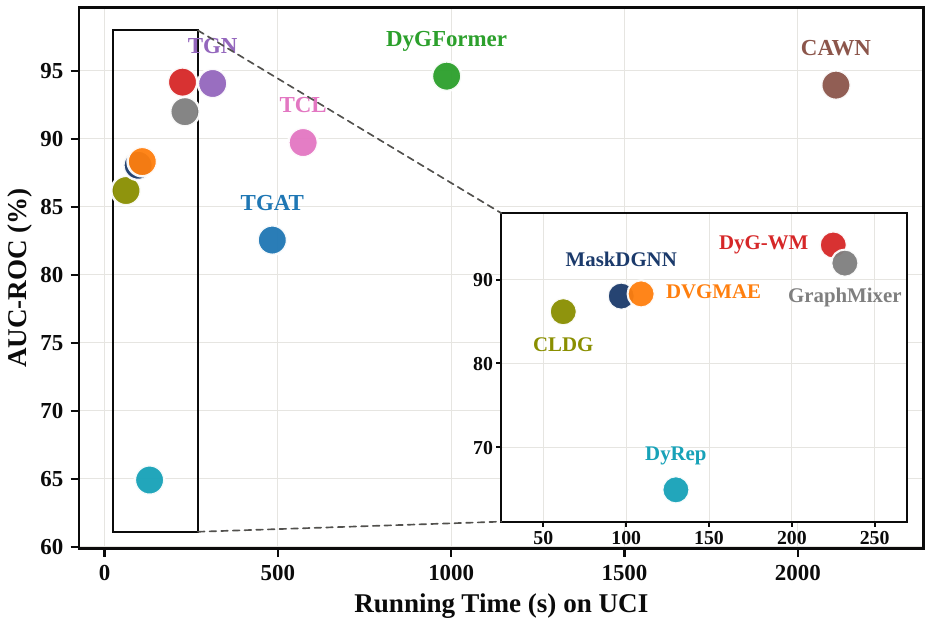}
\vspace{-3pt}
{\scriptsize (f) UCI}
\end{minipage}
\hfill
\begin{minipage}[t]{0.24\textwidth}
\centering
\includegraphics[width=\linewidth]{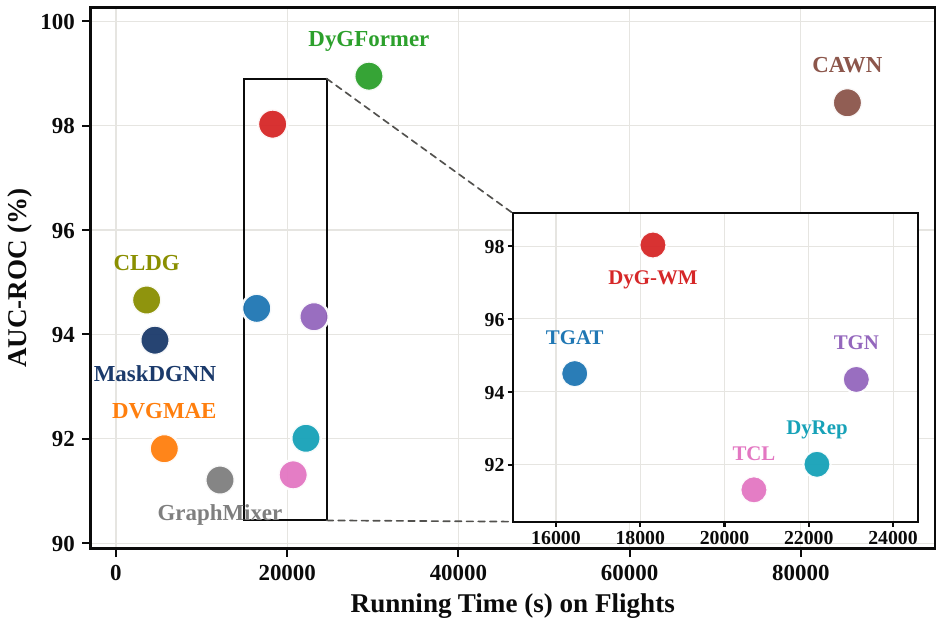}
\vspace{-3pt}
{\scriptsize (g) Flights}
\end{minipage}
\hfill
\begin{minipage}[t]{0.24\textwidth}
\centering
\includegraphics[width=\linewidth]{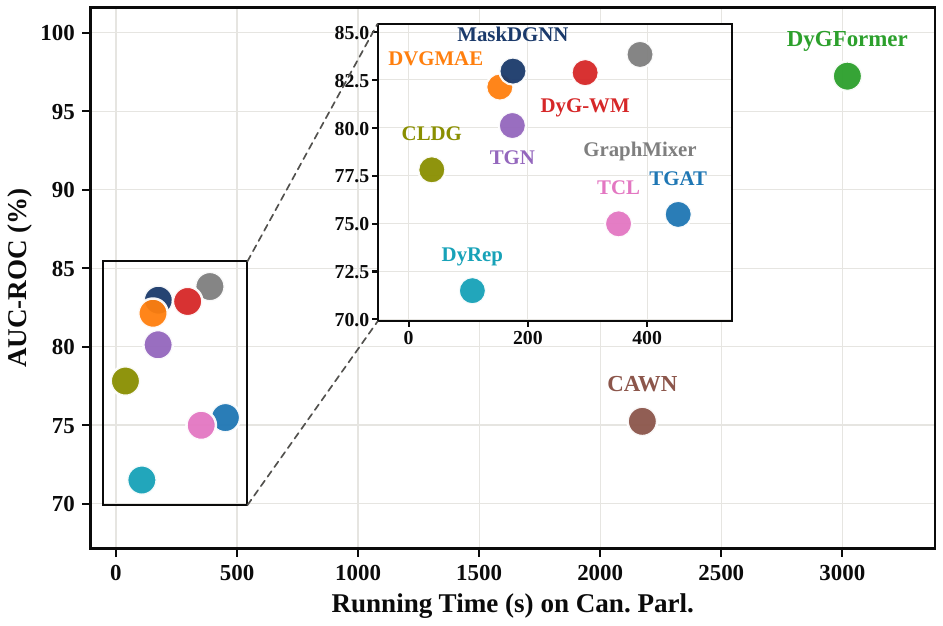}
\vspace{-3pt}
{\scriptsize (h) Can.\ Parl.}
\end{minipage}
\vspace{2mm}
\begin{minipage}[t]{0.24\textwidth}
\centering
\includegraphics[width=\linewidth]{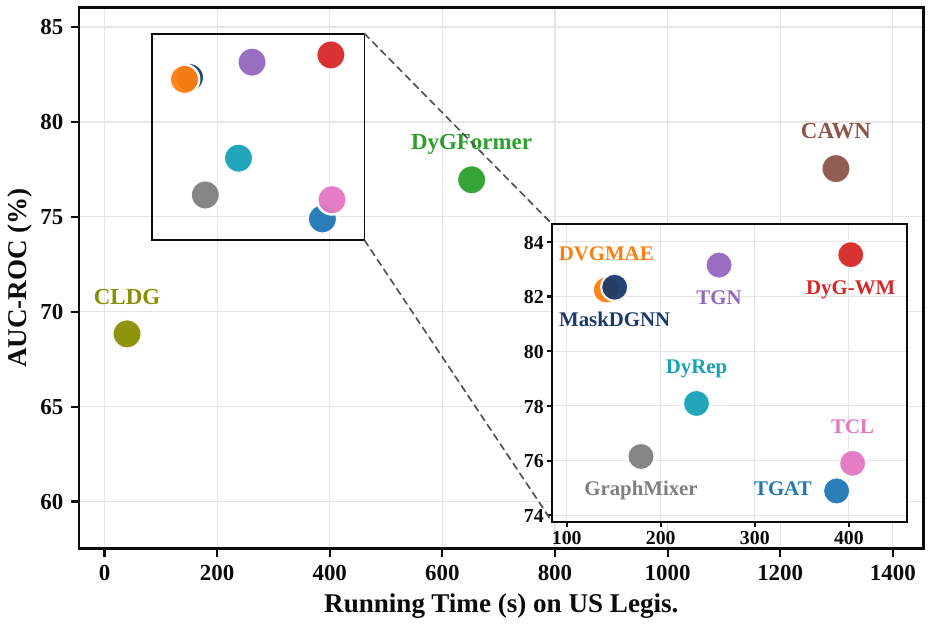}
\vspace{-3pt}
{\scriptsize (i) US Legis.}
\end{minipage}
\hfill
\begin{minipage}[t]{0.24\textwidth}
\centering
\includegraphics[width=\linewidth]{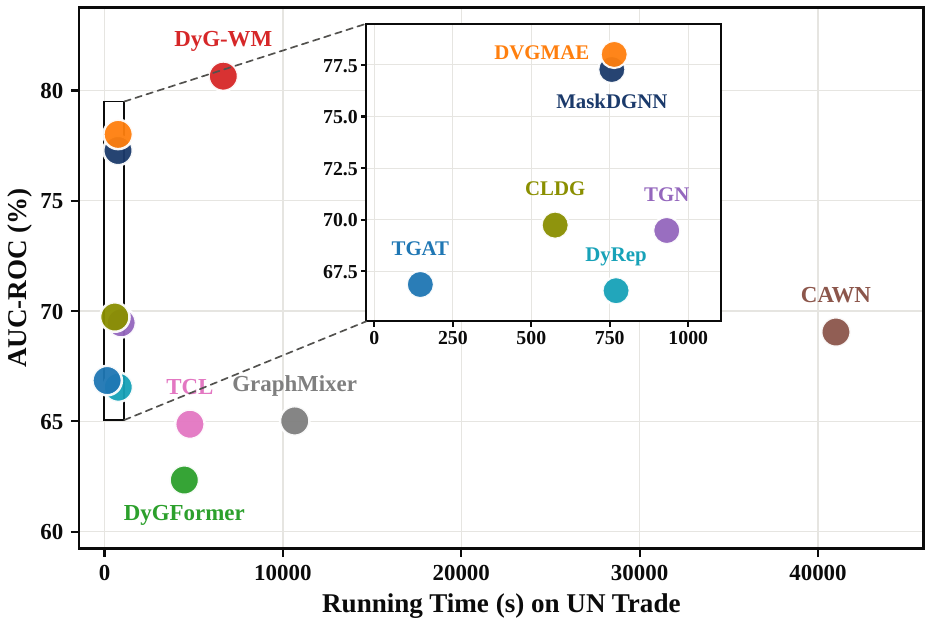}
\vspace{-3pt}
{\scriptsize (j) UN Trade}
\end{minipage}
\hfill
\begin{minipage}[t]{0.24\textwidth}
\centering
\includegraphics[width=\linewidth]{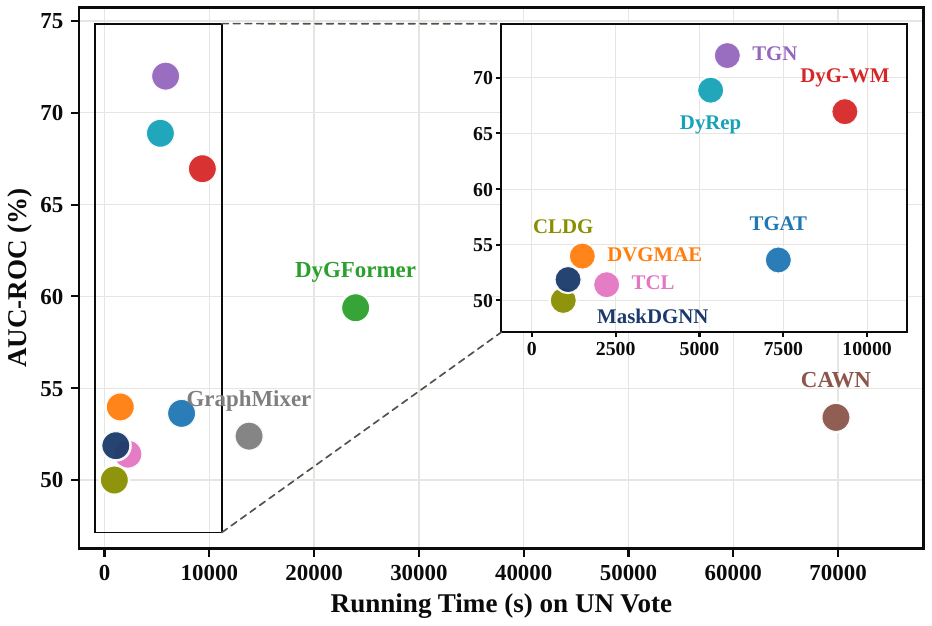}
\vspace{-3pt}
{\scriptsize (k) UN Vote}
\end{minipage}
\hfill
\begin{minipage}[t]{0.24\textwidth}
\centering
\includegraphics[width=\linewidth]{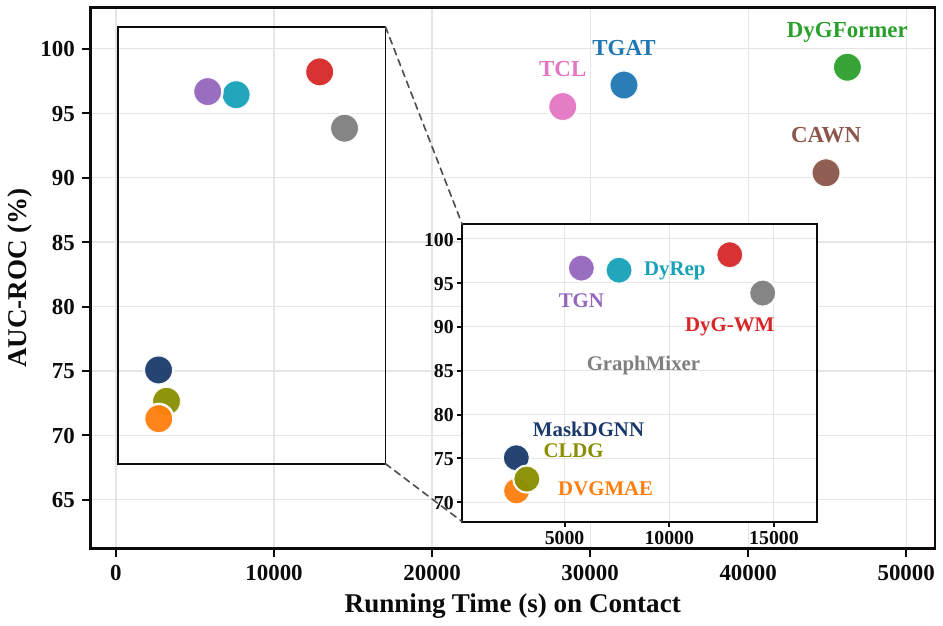}
\vspace{-3pt}
{\scriptsize (l) Contact}
\end{minipage}
\caption{Accuracy-efficiency trade-off across all dynamic link prediction datasets. Each panel reports AUC-ROC against training time.}
\label{fig:efficiency_all}
\end{figure*}

\subsection{Algorithm Details}
This section summarizes \method's representation construction, training, and inference. Algorithm~\ref{alg:historical-representation} constructs representations, Algorithm~\ref{alg:training} gives dual-level predictive training, and Algorithm~\ref{alg:causal-inference} gives link inference without the target branch.

\subsection{Historical Representation Construction}
\label{app:algorithm-context}

For target bin $t$, the online branch uses
$
\mathcal C_t=(\widetilde G_{t-W+1},\ldots,\widetilde G_{t-1}).
$
Each query $(u,v,\tau)$ additionally uses the strict event prefix $\mathcal H_{<\tau}$. Bin-level representations depend only on $\mathcal C_t$, while exact-time representations also depend on $\mathcal H_{<\tau}$.

\begin{algorithm}[t]
\caption{Historical node-relation representation construction}
\label{alg:historical-representation}
\begin{algorithmic}[1]
\REQUIRE Historical snapshots $\mathcal C_t$; queries $\mathfrak P=\{(u_j,v_j,\tau_j)\}_{j=1}^{|\mathfrak P|}$; strict prefixes $\{\mathcal H_{<\tau_j}\}$; online parameters $\Theta$; context budget $B$
\ENSURE Historical and exact-time relation representations

\FOR{$s\in\mathcal I_t$}
    \STATE $\mathbf H^s\leftarrow \operatorname{GNN}_{\theta}(\widetilde G_s)$ and $\mathbf h_x^s\leftarrow\mathbf H^s[x]$ for $x\in\mathcal V$.
\ENDFOR

\FORALL{$x\in\mathcal V$}
    \STATE Encode $\{\mathbf h_x^s\}_{s\in\mathcal I_t}$ and neighborhood trajectories as $\mathbf c_x^t$.
    \STATE Construct $\{\mathbf d_{x,s}^{N}\}_{s\in\mathcal I_t}$ and compute
    $\mathbf S_x^t\leftarrow\phi_N\left(\operatorname{Sig}_2((\mathbf d_{x,s}^{N})_{s\in\mathcal I_t})\right)$.
\ENDFOR

\FOR{$j=1,\ldots,|\mathfrak P|$}
    \STATE $(u,v,\tau)\leftarrow(u_j,v_j,\tau_j)$.
    \STATE Construct $\mathcal P_{uv}^t$ and select
    $\mathcal K_{uv}^t\leftarrow\operatorname{TopB}_{w\in\mathcal P_{uv}^t}\rho_{uv}^t(w)$.
    \FOR{$s\in\mathcal I_t$}
        \STATE Pool selected embeddings into $\mathbf p_{uv}^s$.
    \ENDFOR
    \STATE $\mathbf g_{uv}^t\leftarrow\operatorname{GRU}_{\mathrm{rel}}((\mathbf p_{uv}^s)_{s\in\mathcal I_t})$.
    \STATE Construct $\{\mathbf d_{uv,s}^{R}\}_{s\in\mathcal I_t}$ and compute
    $\mathbf S_{uv}^t\leftarrow\phi_R\left(\operatorname{Sig}_2((\mathbf d_{uv,s}^{R})_{s\in\mathcal I_t})\right)$.
    \STATE $\mathbf r_{uv}^t\leftarrow g_{\theta}\left(\chi(\mathbf c_u^t,\mathbf c_v^t)\right)$.
    \STATE $\mathbf q_{uv}^t\leftarrow\phi_{\mathrm{fuse}}\left([\mathbf r_{uv}^t\Vert\mathbf g_{uv}^t\Vert\mathbf S_{uv}^t]\right)$.
    \STATE $\mathbf o_{uv}^{<\tau}\leftarrow\operatorname{EventHistory}(u,v;\mathcal H_{<\tau})$.
    \STATE $\mathbf z_{uv}^{\tau}\leftarrow\phi_{\tau}\left([\mathbf q_{uv}^t\Vert\mathbf o_{uv}^{<\tau}]\right)$.
\ENDFOR

\RETURN $\{\mathbf c_x^t,\mathbf S_x^t\}_{x\in\mathcal V}$ and $\{\mathbf g_{u_jv_j}^t,\mathbf S_{u_jv_j}^t,\mathbf q_{u_jv_j}^t,\mathbf o_{u_jv_j}^{<\tau_j},\mathbf z_{u_jv_j}^{\tau_j}\}_{j=1}^{|\mathfrak P|}$
\end{algorithmic}
\end{algorithm}

The quantities $\mathbf c_x^t$, $\mathbf g_{uv}^t$, $\mathbf S_x^t$, $\mathbf S_{uv}^t$, and $\mathbf q_{uv}^t$ depend only on $\mathcal C_t$ and are fixed within bin $t$, whereas $\mathbf o_{uv}^{<\tau}$ and $\mathbf z_{uv}^{\tau}$ vary with $\tau$. Events at timestamp $\tau$ are excluded from $\mathcal H_{<\tau}$.

\subsection{Dual-Level Predictive Training}
\label{app:algorithm-training}

Let $\Theta$ contain all trainable online parameters, while $\bar{\theta}$ denotes the EMA target-encoder parameters. Candidate groups remain intact within minibatches so each observed interaction is trained with all sampled negatives.

\begin{algorithm}[t]
\caption{Dual-level joint-embedding predictive training}
\label{alg:training}
\begin{algorithmic}[1]
\REQUIRE Training windows $\{(\mathcal C_t,\widetilde G_t)\}$; event stream $\mathcal E$; negative count $K$; online parameters $\Theta$; loss weights $\lambda_N,\lambda_R,\lambda_{\mathrm{var}},\lambda_{\mathrm{cov}},\lambda_{\mathrm{link}},\lambda_{\mathrm{rank}}$; EMA coefficient $\eta_{\mathrm{ema}}$
\ENSURE Trained online parameters $\Theta$

\STATE Initialize $\bar{\theta}\leftarrow\theta$ and construct the event-history index.

\FOR{each training epoch}
    \FOR{each target bin $t$}
        \STATE Construct candidate groups $\{\mathcal Q(u,\tau)\}$ and partition complete groups into minibatches.
        \FOR{each group-preserving minibatch $\mathfrak P$}
            \STATE Apply Algorithm~\ref{alg:historical-representation}.
            \STATE $\mathbf y_x^t\leftarrow\operatorname{sg}\left(\operatorname{GNN}_{\bar{\theta}}(\widetilde G_t)[x]\right)$ for $x\in\mathcal V_t$.
            \STATE Let $\mathcal R_{t,\mathfrak P}^{+}$ be the observed relation pairs in $\mathfrak P$ and $\mathcal U_t$ their unique endpoints.

            \FORALL{$x\in\mathcal U_t$}
                \STATE $\widehat{\mathbf y}_x^t\leftarrow p_N\left([\mathbf c_x^t\Vert\mathbf S_x^t\Vert\mathbf e_{\Delta_t}]\right)$.
            \ENDFOR

            \FORALL{$(u,v)\in\mathcal R_{t,\mathfrak P}^{+}$}
                \STATE $\widehat{\mathbf y}_{uv}^t\leftarrow p_R\left([\mathbf q_{uv}^t\Vert\mathbf e_{\Delta_t}]\right)$.
                \STATE $\mathbf y_{uv}^t\leftarrow\operatorname{sg}\left(g_{\bar{\theta}}\left(\chi(\mathbf y_u^t,\mathbf y_v^t)\right)\right)$.
            \ENDFOR

            \FORALL{$(u_j,v_j,\tau_j)\in\mathfrak P$}
                \STATE $s_{u_jv_j}^{\tau_j}\leftarrow h_{\psi}(\mathbf z_{u_jv_j}^{\tau_j})$.
            \ENDFOR

            \STATE $\mathcal L^{N}=\frac{1}{|\mathcal U_t|}\sum_{x\in\mathcal U_t}D(\widehat{\mathbf y}_x^t,\mathbf y_x^t)$.
            \STATE $\mathcal L^{R}=\frac{1}{|\mathcal R_{t,\mathfrak P}^{+}|}\sum_{(u,v)\in\mathcal R_{t,\mathfrak P}^{+}}D(\widehat{\mathbf y}_{uv}^t,\mathbf y_{uv}^t)$.
            \STATE Compute node- and relation-level variance and covariance penalties.
            \STATE Compute $\mathcal L_{\mathrm{link}}$ over all candidate scores and $\mathcal L_{\mathrm{rank}}$ within each candidate group.
            \STATE
            \begin{equation}
            \begin{aligned}
            \mathcal L_{\mathrm{JEPA}}={}&\lambda_N\mathcal L^{N} + \lambda_R\mathcal L^{R}\\
            &+\lambda_{\mathrm{var}}\left(\mathcal L_{\mathrm{var}}^{N}+\mathcal L_{\mathrm{var}}^{R}\right)+\lambda_{\mathrm{cov}}\left(\mathcal L_{\mathrm{cov}}^{N}+\mathcal L_{\mathrm{cov}}^{R}\right).
            \end{aligned}
            \label{eq:algorithm-jepa-objective}
            \end{equation}
            \STATE $\mathcal L_{\mathrm{total}}=\mathcal L_{\mathrm{JEPA}}+\lambda_{\mathrm{link}}\mathcal L_{\mathrm{link}}+\lambda_{\mathrm{rank}}\mathcal L_{\mathrm{rank}}$.
            \STATE Update $\Theta$, then $\bar{\theta}\leftarrow\eta_{\mathrm{ema}}\bar{\theta}+(1-\eta_{\mathrm{ema}})\theta$.
        \ENDFOR
    \ENDFOR
\ENDFOR

\RETURN $\Theta$
\end{algorithmic}
\end{algorithm}

Node and relation predictive targets use only observed interactions; sampled negatives enter only $\mathcal L_{\mathrm{link}}$ and $\mathcal L_{\mathrm{rank}}$. Relation targets use target endpoint latents without explicit target-snapshot pair statistics.

For the node benchmark, the relation predictor, event-history branch, link scorer, and relation loss are disabled. The node predictor maps $[\mathbf c_x^t\Vert\mathbf S_x^t\Vert\mathbf e_{\Delta_t}]$ to $\widehat{\mathbf y}_x^t$ and is trained against the target latent $\mathbf y_x^t$ together with node discrimination and anti-collapse terms. Node and link prediction are trained separately.

\subsection{Link Inference}
\label{app:algorithm-inference}

At inference, target encoders and predictive losses are removed. For query $(u,v,\tau)$ in bin $t$, only $\mathcal C_t$ and $\mathcal H_{<\tau}$ are used.

\begin{algorithm}[t]
\caption{Dynamic link inference}
\label{alg:causal-inference}
\begin{algorithmic}[1]
\REQUIRE Historical snapshots $\mathcal C_t$; candidate queries $\mathfrak P=\{(u_j,v_j,\tau_j)\}$; strict event-history index; trained online parameters $\Theta$
\ENSURE Interaction scores $\{s_{u_jv_j}^{\tau_j}\}_{j=1}^{|\mathfrak P|}$

\STATE Encode only $\mathcal C_t$; do not invoke $\operatorname{GNN}_{\bar{\theta}}$, $g_{\bar{\theta}}$, or encode $\widetilde G_t$.

\FOR{$j=1,\ldots,|\mathfrak P|$}
    \STATE $(u,v,\tau)\leftarrow(u_j,v_j,\tau_j)$.
    \STATE Construct $\mathbf c_u^t,\mathbf c_v^t,\mathbf g_{uv}^t,\mathbf S_{uv}^t,\mathbf q_{uv}^t$ from $\mathcal C_t$.
    \STATE Query the event-history index for events satisfying $\tau_i<\tau$ to obtain $\mathbf o_{uv}^{<\tau}$.
    \STATE $\mathbf z_{uv}^{\tau}\leftarrow\phi_{\tau}\left([\mathbf q_{uv}^{t}\Vert\mathbf o_{uv}^{<\tau}]\right)$.
    \STATE $s_{uv}^{\tau}\leftarrow h_{\psi}(\mathbf z_{uv}^{\tau})$.
\ENDFOR

\RETURN $\{s_{u_jv_j}^{\tau_j}\}_{j=1}^{|\mathfrak P|}$
\end{algorithmic}
\end{algorithm}

Thus,
$s_{uv}^{\tau}=\operatorname{GNN}_{\theta}(u,v,\tau;\mathcal C_t,\mathcal H_{<\tau})$
and neither $\widetilde G_t$ nor the target latents $\mathbf y_u^t,\mathbf y_{uv}^t$ enter inference.

\section{Computational Complexity}
\label{app:complexity}

We analyze one target bin $t$ with $\mathcal I_t=\{t-W+1,\ldots,t-1\}$ and $L=|\mathcal I_t|=W-1$. Let
$m_s=\|A_s\|_0, M_{\mathcal C}=\sum_{s\in\mathcal I_t}m_s$,
where $m_s$ and $M_{\mathcal C}$ are the numbers of stored message-passing edges in snapshot $s$ and the historical context, respectively. Let $N=|\mathcal V|$, $d$ be the hidden dimension, $d_s$ the projected path dimension, $B$ the relation-context budget, and $P=|\mathfrak P|$ the number of candidate rows in a group-preserving minibatch. With $G$ observed interactions and $K$ negatives per interaction, $P=G(K+1)$.

Historical snapshot encodings and node contexts are shared across candidates, while relation contexts, relation signatures, and event-history features are pair-conditioned.

\subsection{Historical Snapshot and Node Encoding}

For a fixed-depth sparse message-passing implementation of $\operatorname{GNN}_{\theta}$, encoding all historical snapshots costs
\begin{equation}
\mathcal T_{\mathrm{snap}}=\mathcal O\left(M_{\mathcal C}d+LNd^2\right).
\label{eq:complexity-snapshot}
\end{equation}
No dense $N\times N$ adjacency matrix is materialized.

Sparse neighborhood pooling costs $\mathcal O(M_{\mathcal C}d)$, node-wise recurrent updates cost $\mathcal O(LNd^2)$, and node-context fusion costs $\mathcal O(Nd^2)$. Hence,
\begin{equation}
\mathcal T_{\mathrm{node}}=\mathcal O\left(M_{\mathcal C}d+LNd^2\right).
\label{eq:complexity-node}
\end{equation}
Both terms are shared across all $P$ candidate rows.

\subsection{Relation-Centric Context Selection}

For candidate pair $(u_j,v_j)$, define
\begin{equation}
D_j=\sum_{s\in\mathcal I_t}\left(|\mathcal N_s(u_j)|+|\mathcal N_s(v_j)|\right).
\label{eq:pair-neighborhood-volume}
\end{equation}
The number of distinct candidate context nodes is
\begin{equation}
C_j=\left|\{u_j,v_j\}\cup\bigcup_{s\in\mathcal I_t}\left(\mathcal N_s(u_j)\cup\mathcal N_s(v_j)\right)\right|.
\label{eq:pair-candidate-count}
\end{equation}
Constructing the candidate union and scores costs $\mathcal O(D_j)$, while sorting costs $\mathcal O(C_j\log C_j)$. Thus,
\begin{equation}
\mathcal T_{\mathrm{select}}=\mathcal O\left(D_{\mathfrak P}+\sum_{j=1}^{P}C_j\log C_j\right),\qquad D_{\mathfrak P}=\sum_{j=1}^{P}D_j.
\label{eq:complexity-selection}
\end{equation}
Partial top-$B$ selection can reduce the sorting term to $\mathcal O(C_j\log B)$.

Pooling at most $B$ embeddings over $L$ snapshots costs $\mathcal O(PLBd)$, and the relation GRU costs $\mathcal O(PLd^2)$. Therefore,
\begin{equation}
\mathcal T_{\mathrm{relctx}}=\mathcal O\left(D_{\mathfrak P}+\sum_{j=1}^{P}C_j\log C_j+PLBd+PLd^2\right).
\label{eq:complexity-relation-context}
\end{equation}

\subsection{Node-Relation Path Signatures}

For $L$ increments in $\mathbb R^{d_s}$, first-order accumulation costs $\mathcal O(Ld_s)$ and the second-order signature costs $\mathcal O(Ld_s^2)$ time and $\mathcal O(d_s^2)$ working memory.

For all $N$ nodes,
\begin{equation}
\mathcal T_{\mathrm{sig}}^{N}=\mathcal O\left(M_{\mathcal C}+NLd_s^2+Nd_s^2d\right).
\label{eq:complexity-node-signature}
\end{equation}
The terms correspond to structural-transition extraction, second-order signature updates, and projection to dimension $d$.

For $P$ candidate relations,
\begin{equation}
\mathcal T_{\mathrm{sig}}^{R}=\mathcal O\left(D_{\mathfrak P}+PLd_s^2+Pd_s^2d\right).
\label{eq:complexity-relation-signature}
\end{equation}
Node signatures are shared, whereas relation signatures are pair-conditioned.

\subsection{Exact-Time Event-History Queries}

The event stream contains $M$ raw interactions. Timestamp, directed-pair, and endpoint indices are constructed once in $\mathcal O(M\log M)$ time. Let $U$ denote the number of distinct directed pairs. Without semantic profiles, the persistent index requires
\begin{equation}
\mathcal O\left(M+T(U+N)\right)
\label{eq:history-index-memory}
\end{equation}
memory. Enabling $d_m$-dimensional semantic profiles adds $\mathcal O(T(U+N)d_m)$.

For query $(u,v,\tau)$, binary search retrieves the required strict-prefix statistics in $\mathcal O(\log M)$ time. Across one minibatch,
\begin{equation}
\mathcal T_{\mathrm{hist}}=\mathcal O\left(P\log M+Pd^2\right).
\label{eq:complexity-history}
\end{equation}
The second term covers event-history projection and fusion.

\subsection{Prediction Heads and Training Objective}
The endpoint transformation, relation encoder, fusion modules, predictors, and link scorer cost $\mathcal O(Pd^2)$. Pointwise losses cost $\mathcal O(Pd)$ and groupwise ranking at most $\mathcal O(P\log P)$.

Let
$P_+=|\mathcal R_{t,\mathfrak P}^{+}|, U_+=|\mathcal U_t|$,
where $P_+$ is the number of observed relation pairs and $U_+$ their unique endpoints. Since $U_+\leq2P_+$, the node- and relation-level variance terms cost $\mathcal O(P_+d)$, while the covariance terms cost
\begin{equation}
\mathcal T_{\mathrm{cov}}=\mathcal O\left((U_++P_+)d^2\right)=\mathcal O(P_+d^2).
\label{eq:complexity-covariance}
\end{equation}
The covariance term is omitted at a level with fewer than two prediction vectors.
\subsection{Training-Time Overhead}
Training additionally encodes the target snapshot $\widetilde G_t$. Let $m_t=\|A_t\|_0$. Target snapshot encoding and positive relation-target construction cost
\begin{equation}
\mathcal T_{\mathrm{target}}=\mathcal O\left(m_td+Nd^2+P_+d^2\right).
\label{eq:complexity-target}
\end{equation}
Including covariance regularization and the EMA update,
\begin{equation}
\mathcal T_{\mathrm{train}}=\mathcal O\left(\mathcal T_{\mathrm{infer}}+m_td+Nd^2+P_+d^2+|\Theta_{\mathrm{EMA}}|\right).
\label{eq:complexity-training}
\end{equation}
This excludes the constant-factor backward pass through the online branch. Since $U_+\leq2P_+$, the $P_+d^2$ term covers positive relation-target encoding and both covariance computations. The EMA update is independent of graph size and candidate count.

Each group-preserving minibatch performs one parameter update, so historical and target snapshots are re-encoded after each update. At inference, historical representations for a target bin can be reused across candidate minibatches.

\subsection{Memory Complexity}

Excluding model parameters and the persistent event-history index, inference requires
\begin{equation}
\mathcal S_{\mathrm{infer}}=\mathcal O\left(M_{\mathcal C}+LNd+PB+PLd+(N+P)d_s^2\right).
\label{eq:memory-inference}
\end{equation}
Training additionally stores target embeddings, online activations, covariance matrices, gradients, and optimizer states:
\begin{equation}
\mathcal S_{\mathrm{train}}=\mathcal O\left(\mathcal S_{\mathrm{infer}}+Nd+P_+d+d^2+|\Theta|\right).
\label{eq:memory-training}
\end{equation}

Relation-centric modeling is bounded by $B$, while the second-order signature remains linear in $L$. The target branch is training-only, so inference uses only $\mathcal C_t$ and $\mathcal H_{<\tau}$.
\section{Proofs and Technical Derivations}
\label{app:proofs}

We establish three properties of \method: de-duplicated snapshots do not generally identify pairwise interaction multiplicity, second-order signatures capture temporal order beyond first-order summaries, and the deployed scorer depends only on the historical snapshot context and strict event prefix.

\subsection{Non-Identifiability of Pairwise Interaction Multiplicity}
\label{app:proof-nonidentifiability}

For a directed pair $(u,v)$, define its multiplicity in bin $t$ as
\begin{equation}
n_{uv}^{t}=\left|\left\{(u_i,v_i,\tau_i,\mathbf e_i)\in\mathcal E_t:u_i=u,\ v_i=v\right\}\right|.
\label{eq:pair-multiplicity}
\end{equation}

For the following proposition, assume that $X_t$ contains fixed node attributes and nodewise event aggregates, such as incident interaction counts, nodewise link-feature aggregates, and last-activity times, but no pair-resolved statistic such as $n_{uv}^{t}$.

\noindent \textbf{Proposition} 1 (Snapshot non-identifiability).

Under this snapshot construction, the mapping from the raw event stream to the snapshot sequence is not injective. For every integer $k\geq1$, there exist two streams $\mathcal S_k$ and $\mathcal S'_k$ with identical snapshot views $\widetilde G_s=(\mathcal V,A_s,X_s,\mathbf m_s)$ for every bin $s$, but for some directed pair,
$\left|n_{uv}^{t}-n_{uv}^{\prime,t}\right|=k$.

\noindent \textbf{Proof.}
Consider four distinct nodes $a,b,c,d$, with $a,b$ as sources and $c,d$ as destinations. The two streams are identical outside bin $t$. Within bin $t$, all interactions share timestamp $\tau_0$ and feature $\mathbf e_0$, with multiplicities
\begin{equation}
\begin{array}{c|cc}
\mathcal S_k & c & d\\
\hline
a & k+1 & 1\\
b & 1 & k+1
\end{array}
\qquad
\begin{array}{c|cc}
\mathcal S'_k & c & d\\
\hline
a & 1 & k+1\\
b & k+1 & 1
\end{array}.
\label{eq:counterexample-counts}
\end{equation}

Both streams contain the same directed edge support
$\{(a,c),(a,d),(b,c),(b,d)\}$,
so de-duplication and symmetrization give identical $A_t$. Each source has $k+2$ outgoing interactions and each destination has $k+2$ incoming interactions in both streams. Since all interactions share $\mathbf e_0$ and $\tau_0$, nodewise link-feature aggregates and last-activity statistics are also identical. The active-node masks coincide, and the streams are identical outside bin $t$. Hence,
$\widetilde G_s(\mathcal S_k)=\widetilde G_s(\mathcal S'_k)$.

For pair $(a,c)$,
$n_{ac}^{t}=k+1,  n_{ac}^{\prime,t}=1$,
so
$\left|n_{ac}^{t}-n_{ac}^{\prime,t}\right|=k$.
Since $k$ is arbitrary, the snapshot sequence does not determine pairwise interaction multiplicity.

\noindent\textbf{Consequence.}
For any deterministic encoder $F$ whose graph input is only $\mathcal C_t$,
\begin{equation}
F\bigl(\mathcal C_t(\mathcal S_k)\bigr)=F\bigl(\mathcal C_t(\mathcal S'_k)\bigr),
\label{eq:snapshot-encoder-indistinguishable}
\end{equation}
although the underlying pairwise recurrence patterns differ. We therefore retain $\mathcal H_{<\tau}$ separately to preserve direction and repeated interactions. The proposition does not apply to snapshot representations that explicitly store pairwise multiplicities.

\subsection{Order Sensitivity of the Second-Order Signature}
\label{app:proof-signature}

For projected increments $\mathbf d_1,\ldots,\mathbf d_L\in\mathbb R^{d_s}$, define
\begin{equation}
\operatorname{Sig}^{(1)}=\sum_{i=1}^{L}\mathbf d_i,
\label{eq:proof-signature-first}
\end{equation}
and
\begin{equation}
\operatorname{Sig}^{(2)}=\frac{1}{2}\sum_{i=1}^{L}\mathbf d_i\otimes\mathbf d_i+\sum_{i<j}\mathbf d_i\otimes\mathbf d_j.
\label{eq:proof-signature-second}
\end{equation}

\noindent \textbf{Proposition} 2 (Two-step order sensitivity).

For $\mathbf a,\mathbf b\in\mathbb R^{d_s}$, the paths $(\mathbf a,\mathbf b)$ and $(\mathbf b,\mathbf a)$ have identical first-order signatures, while
\begin{equation}
\operatorname{Sig}^{(2)}(\mathbf a,\mathbf b)-\operatorname{Sig}^{(2)}(\mathbf b,\mathbf a)=\mathbf a\otimes\mathbf b-\mathbf b\otimes\mathbf a.
\label{eq:signature-order-difference}
\end{equation}
If both increments are nonzero and $\vartheta$ is their angle,
\begin{equation}
\left\|\operatorname{Sig}^{(2)}(\mathbf a,\mathbf b)-\operatorname{Sig}^{(2)}(\mathbf b,\mathbf a)\right\|_F=\sqrt{2}\,\|\mathbf a\|_2\|\mathbf b\|_2|\sin\vartheta|.
\label{eq:signature-order-gap}
\end{equation}
Thus, the second-order signature distinguishes the two orderings exactly when the increments are not collinear.

\noindent \textbf{Proof.}
The first-order signatures satisfy
$\operatorname{Sig}^{(1)}(\mathbf a,\mathbf b)=\mathbf a+\mathbf b=\operatorname{Sig}^{(1)}(\mathbf b,\mathbf a)$.
From Eq.~\ref{eq:proof-signature-second},
\begin{equation}
\begin{aligned}
\operatorname{Sig}^{(2)}(\mathbf a,\mathbf b)&=\frac{1}{2}\mathbf a\otimes\mathbf a+\mathbf a\otimes\mathbf b+\frac{1}{2}\mathbf b\otimes\mathbf b,\\
\operatorname{Sig}^{(2)}(\mathbf b,\mathbf a)&=\frac{1}{2}\mathbf b\otimes\mathbf b+\mathbf b\otimes\mathbf a+\frac{1}{2}\mathbf a\otimes\mathbf a.
\end{aligned}
\end{equation}
Subtracting gives Eq.~\ref{eq:signature-order-difference}. Let
$\mathbf M=\mathbf a\mathbf b^\top-\mathbf b\mathbf a^\top$.
Then
\begin{equation}
\begin{aligned}
\|\mathbf M\|_F^2&=2\|\mathbf a\|_2^2\|\mathbf b\|_2^2-2(\mathbf a^\top\mathbf b)^2\\
&=2\|\mathbf a\|_2^2\|\mathbf b\|_2^2\sin^2\vartheta,
\end{aligned}
\end{equation}
which gives Eq.~\ref{eq:signature-order-gap}.

\noindent\textbf{General Decomposition.}
For arbitrary $L$, define
\begin{equation}
\mathbf A=\frac{1}{2}\left(\operatorname{Sig}^{(2)}-\operatorname{Sig}^{(2)\top}\right).
\label{eq:signature-antisymmetric}
\end{equation}
Then
\begin{equation}
\mathbf A=\frac{1}{2}\sum_{i<j}\left(\mathbf d_i\otimes\mathbf d_j-\mathbf d_j\otimes\mathbf d_i\right),
\label{eq:signature-levy-area}
\end{equation}
while
\begin{equation}
\frac{1}{2}\left(\operatorname{Sig}^{(2)}+\operatorname{Sig}^{(2)\top}\right)=\frac{1}{2}\operatorname{Sig}^{(1)}\otimes\operatorname{Sig}^{(1)}.
\label{eq:signature-symmetric}
\end{equation}
Hence,
\begin{equation}
\operatorname{Sig}^{(2)}=\frac{1}{2}\operatorname{Sig}^{(1)}\otimes\operatorname{Sig}^{(1)}+\mathbf A.
\label{eq:signature-decomposition}
\end{equation}
The symmetric component is determined by the first-order signature, whereas $\mathbf A$ contains additional ordered cross-step information. This establishes order sensitivity, not full injectivity: distinct longer paths may still share the same truncated second-order signature.

\subsection{Prefix of the Deployed Scorer}
\label{app:proof-causality}

The following claim concerns the deployed scorer with fixed learned parameters and an offline snapshot partition; it does not require the learned parameters to be independent of target snapshots used during training.

\noindent \textbf{Proposition} 3 (Prefix scoring invariance).

Fix learned online parameters $\Theta$, node identities, and an offline snapshot partition. Consider two streams $\mathcal S$ and $\mathcal S'$ with the same query $(u,v,\tau)$ in target bin $t$. If
\begin{equation}
\mathcal C_t(\mathcal S)=\mathcal C_t(\mathcal S'),\qquad \mathcal H_{<\tau}(\mathcal S)=\mathcal H_{<\tau}(\mathcal S'),
\label{eq:causal-equivalence-assumption}
\end{equation}
then
\begin{equation}
s_{uv}^{\tau}(\mathcal S;\Theta)=s_{uv}^{\tau}(\mathcal S';\Theta).
\label{eq:causal-score-invariance}
\end{equation}

\noindent \textbf{Proof.}
At inference, $\widetilde G_t$, $\operatorname{GNN}_{\bar{\theta}}$, and $g_{\bar{\theta}}$ are not used, while the online encoders remain part of the scorer. Equality of $\mathcal C_t$ gives identical historical representations:
\begin{equation}
\begin{aligned}
\mathbf c_u^t(\mathcal S)&=\mathbf c_u^t(\mathcal S'),&
\mathbf c_v^t(\mathcal S)&=\mathbf c_v^t(\mathcal S'),\\
\mathbf g_{uv}^t(\mathcal S)&=\mathbf g_{uv}^t(\mathcal S'),&
\mathbf S_{uv}^t(\mathcal S)&=\mathbf S_{uv}^t(\mathcal S').
\end{aligned}
\end{equation}
Since
\begin{equation}
\mathbf q_{uv}^t=\phi_{\mathrm{fuse}}\left([g_{\theta}(\chi(\mathbf c_u^t,\mathbf c_v^t))\Vert\mathbf g_{uv}^t\Vert\mathbf S_{uv}^t]\right),
\label{eq:proof-historical-relation}
\end{equation}
we have
$\mathbf q_{uv}^t(\mathcal S)=\mathbf q_{uv}^t(\mathcal S')$.
Equality of the strict prefixes similarly gives
$\mathbf o_{uv}^{<\tau}(\mathcal S)=\mathbf o_{uv}^{<\tau}(\mathcal S')$.
Therefore,
\begin{equation}
\begin{aligned}
\mathbf z_{uv}^{\tau}(\mathcal S)
&=\phi_{\tau}\left([\mathbf q_{uv}^t(\mathcal S)\Vert\mathbf o_{uv}^{<\tau}(\mathcal S)]\right)\\
&=\phi_{\tau}\left([\mathbf q_{uv}^t(\mathcal S')\Vert\mathbf o_{uv}^{<\tau}(\mathcal S')]\right)\\
&=\mathbf z_{uv}^{\tau}(\mathcal S').
\end{aligned}
\end{equation}
Since
$s_{uv}^{\tau}=h_{\psi}\left(\mathbf z_{uv}^{\tau}\right)$,
the scores are identical.

\noindent\textbf{Scope of the Claim.}
Proposition~3 assumes a fixed offline partition. Given this partition, deployed scoring uses only $\mathcal C_t$ and $\mathcal H_{<\tau}$.

The claim does not require $\mathbf y_{uv}^t$ to be independent of the queried interaction. Although $g_{\bar{\theta}}$ receives no explicit target-bin pair counts or pair-resolved event features, its endpoint latents are computed from $\widetilde G_t$ and may depend indirectly on its adjacency. Target-bin information may therefore affect learned parameters during training, but $\widetilde G_t$, $\mathbf y_u^t$, and $\mathbf y_{uv}^t$ are not scoring inputs at inference.
\end{document}